\def\paperlanguage{}
\pdfoutput=1 

\documentclass[letterpaper, 10 pt, conference]{ieeeconf}  

\usepackage{bm}
\usepackage{cite}
\usepackage{flushend}
\include{preamble}

\IEEEoverridecommandlockouts                              

\title{\LARGE \textbf
  {
    \switchlanguage%
    {%
      Toward Autonomous Driving by Musculoskeletal Humanoids: A Study of Developed Hardware and Learning-Based Software
    }%
    {%
      筋骨格ヒューマノイドによる自動運転に向けて
    }%
  }
}

\author{Kento Kawaharazuka$^{1}$, Kei Tsuzuki$^{1}$, Yuya Koga$^{1}$, Yusuke Omura$^{1}$, Tasuku Makabe$^{1}$, Koki Shinjo$^{1}$\\Moritaka Onitsuka$^{1}$, Yuya Nagamatsu$^{1}$, Yuki Asano$^{1}$, Kei Okada$^{1}$, Koji Kawasaki$^{2}$,  and Masayuki Inaba$^{1}$
  \thanks{$^{1}$ The authors are with the Department of Mechano-Informatics, Graduate School of Information Science and Technology, The University of Tokyo, 7-3-1 Hongo, Bunkyo-ku, Tokyo, 113-8656, Japan.
    {\texttt\small kawaharazuka@jsk.t.u-tokyo.ac.jp}
  }
  \thanks{$^{2}$ The author is associated with TOYOTA MOTOR CORPORATION.
  }
}

\begin{document}

\maketitle
\thispagestyle{empty}
\pagestyle{empty}

\begin{abstract}
  \switchlanguage%
  {%
    This paper summarizes an autonomous driving project by musculoskeletal humanoids.
    The musculoskeletal humanoid, which mimics the human body in detail, has redundant sensors and a flexible body structure.
    These characteristics are suitable for motions with complex environmental contact, and the robot is expected to sit down on the car seat, step on the acceleration and brake pedals, and operate the steering wheel by both arms.
    We reconsider the developed hardware and software of the musculoskeletal humanoid Musashi in the context of autonomous driving.
    The respective components of autonomous driving are conducted using the benefits of the hardware and software.
    Finally, Musashi succeeded in the pedal and steering wheel operations with recognition.
  }%
  {%
    本稿は, 筋骨格ヒューマノイドによる自動運転プロジェクトの要約である.
    人体を詳細に模倣した筋骨格ヒューマノイドは冗長なセンサと柔軟な身体を有し, これを用いて人間同様に車を運転できると考えられる.
    本稿ではそのモジュラー型で冗長なセンサを有し柔軟なハードウェア構成, それらのセンサ・身体を扱うための学習型ソフトウェアについて述べる.
    それら一つ一つの構成要素の自動運転に関する利点について述べる.
    最後に, 筋骨格ヒューマノイドMusashiにより, 認識を含むペダル操作・ハンドル操作実験を行う.
  }%
\end{abstract}

\section{INTRODUCTION}\label{sec:introduction}
\begin{figure}[t]
  \centering
  \includegraphics[width=0.8\columnwidth]{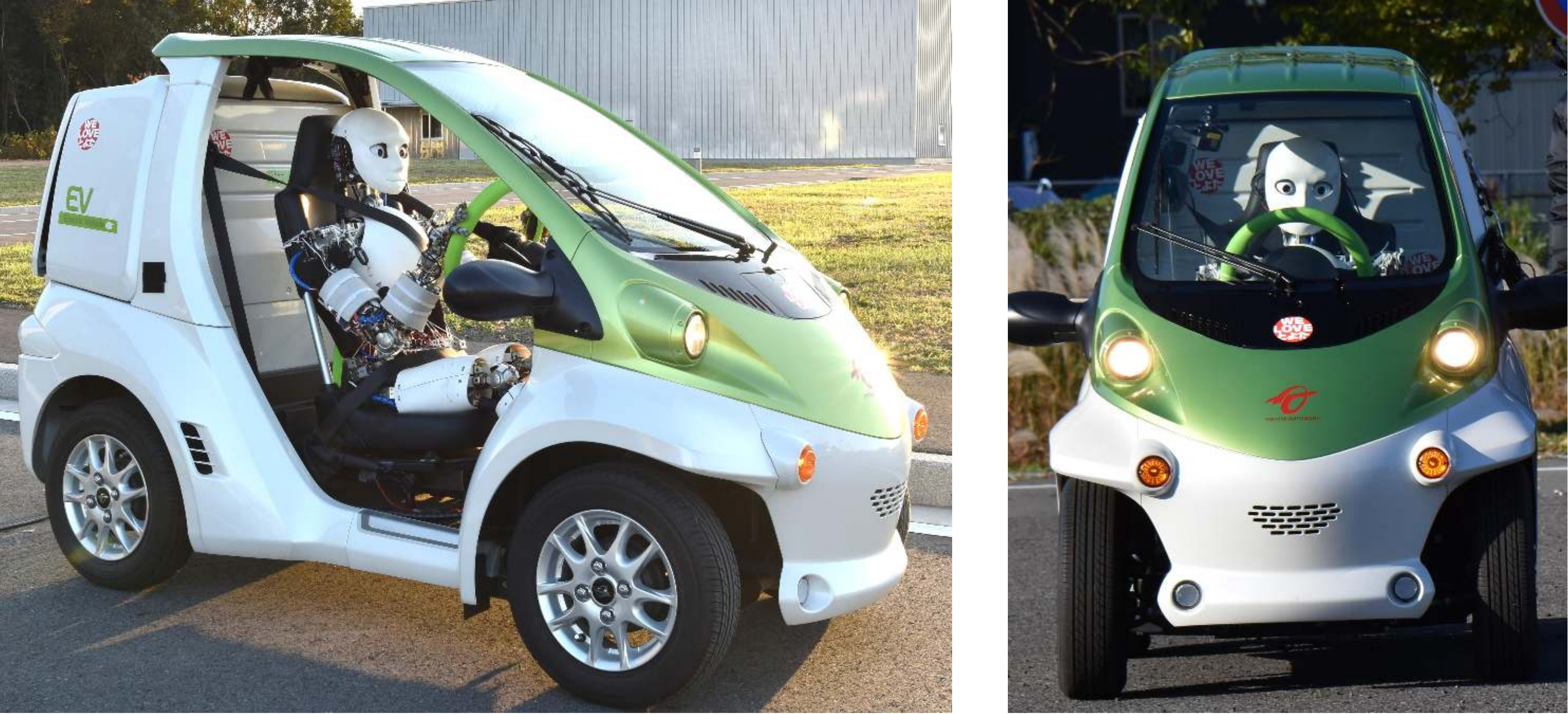}
  \caption{Autonomous driving by musculoskeletal humanoids}
  \label{figure:motivation}
  \vspace{-3.0ex}
\end{figure}

\switchlanguage%
{%
  As a means of safe and comfortable transportation, various researches in autonomous driving are in progress \cite{levinson2011driving, endsley2017tesla}.
  Some companies have achieved a certain level of autonomous driving, e.g. headway vehicle following and autonomous parking.
  These cars are equipped with powerful cameras, LiDARs, GPS, and processors for accurate and safe control.

  On the other hand, starting with the autonomous driving task at DARPA Robotics Challenge (DRC) \cite{darpa2015drc}, there are researches on autonomous driving by humanoid robots \cite{rasmussen2014driving, paolillo2018driving}.
  Humanoid robots are equipped with various sensors for visual, acoustic, and force information.
  By using these sensors, the robot can get into the car and drive instead of humans.
  Also, unlike ordinary autonomous driving, the humanoid robot is expected to do other various tasks, e.g. carrying heavy baggage, assisting the old, doing housework, and aiding with disaster response.
  However, because the humanoid robot lacks the flexibility of the body and deviates from the human body proportion, in DRC, a steering wheel was operated by one arm and a special jig was required to sit down on the seat.

  Although various kinds of humanoid robots exist, one that imitates the human actuation system is the musculoskeletal humanoid \cite{gravato2010ecce1, nakanishi2013design, asano2016kengoro, kawaharazuka2019musashi}.
  The musculoskeletal humanoid is actuated not by motors arranged at each axis but by pneumatic actuators or muscle actuators with motors that imitate human muscles.
  The body is flexible compared with the ordinary axis-driven humanoid due to the muscle elasticity and under-actuation, and so it is suitable for motions with complex environmental contact.
  Therefore, the musculoskeletal humanoid can sit down on the car seat easily without a special jig and operate the steering wheel flexibly by both arms.
  Also, this robot can be used as a more realistic crash test dummy \cite{haug2004crash} because of the human-like body structure and actuation.

  In this study, we introduce our project of autonomous driving by musculoskeletal humanoids (\figref{figure:motivation}).
  Especially, we describe the characteristics of the hardware and learning-based software to move the flexible body.
  We reconsider the developed hardware \cite{kawaharazuka2019musashi, makabe2018eyeunit, makino2018hand, shinjo2019foot} and software \cite{kawaharazuka2019longtime, kawaharazuka2019pedal, kawaharazuka2019relax} in the context of autonomous driving.
  Also, we develop the respective components of autonomous driving and conduct experiments integrating them using the hardware and software characteristics.

  This paper is organized as follows.
  In \secref{sec:hardware}, we will explain the characteristics of the developed hardware and their applications to autonomous driving.
  In \secref{sec:software}, we will explain the overview of the learning-based software and their application to autonomous driving.
  In \secref{sec:experiments}, we will integrate the hardware and software, and conduct experiments of the pedal and handle operations with recognition.
  In \secref{sec:discussion}, we will discuss the issues of each experiment and future directions, and finally conclude this project in \secref{sec:conclusion}.
}%
{%
  現在, 安全で快適な移動手段として, 自動運転技術に関する様々な研究開発が行われている\cite{levinson2011driving, endsley2017tesla}.
  実際に複数の企業において, 渋滞時の追従走行や自動駐車等が実現されている.
  これら自動運転車には, 強力なカメラ・レーダー・GPS・プロセッサ等のシステムが搭載されており, これらを使って車を制御する.

  一方で, DARPA Robotics Challenge \cite{darpa2015drc}における車走行タスクを初めとして, ヒューマノイドロボットによる自動運転の研究も存在する\cite{rasmussen2014driving, paolillo2018driving}.
  ヒューマノイドロボットは視覚や聴覚, 力覚等に関する様々なセンサを搭載しており, これらを用いて車に乗り込み, 運転を人間の代わりに行うことができる.
  また, 自動運転車とは違い, ヒューマノイドロボットはその他荷物運びや高齢者のアシスト, 家事や災害対応等への応用が期待されており, 一台で様々なタスクを成し遂げられると考えられている.
  しかし, これらのヒューマノイドは柔軟性に欠け, かつ身体構造が人間からいくらか逸脱しているため, DRCにおいてハンドル操作は全て片手で行っており, 座るのにも特殊なJIGを要する等の問題点があった.

  ヒューマノイドロボットにも様々なタイプが存在するが, その中でもより人間のアクチュエーションシステムを模倣したのが筋骨格ヒューマノイドである\cite{gravato2010ecce1, nakanishi2013design, asano2016kengoro, kawaharazuka2019musashi}.
  筋骨格ヒューマノイドは通常のロボットのような軸ごとに配置されたモータではなく, 筋肉を模した空気圧やモータによる腱駆動アクチュエータを用いて身体が駆動されている.  その身体は軸駆動型ヒューマノイドに比べ, 筋の弾性や劣駆動性による柔軟性を持ち, 環境に身体を接触させて動作を行うのに適している.
  そのため, より容易に人間の使う実際の車に乗り込むことができ, DARPAにおけるロボットの特殊な座り方やそのためのJIG, 片手によるハンドル操作等を解消することができると考える.
  また, 人間同様の身体システムは人体シミュレータとしての活用も考えられ, 自動運転に関する衝撃シミュレータ\cite{haug2004crash}への応用が挙げられる.

  本研究では, 筋骨格ヒューマノイドによる自動運転実現のための我々の取り組みを紹介する(\figref{figure:motivation}).
  特に, 自動運転に着目した筋骨格ヒューマノイドのハードウェア特徴, それらを動作させるための学習型ソフトウェアの特徴について主に述べる.
  これまで開発してきたハードウェア\cite{kawaharazuka2019musashi, makabe2018eyeunit, makino2018hand, shinjo2019foot}・ソフトウェア\cite{kawaharazuka2019longtime, kawaharazuka2019pedal, kawaharazuka2019relax}を自動運転というコンテキストで捉え直し, 認識等を含む全体システムを開発する.
  また, これらのハードウェア・ソフトウェアを用いた自動運転に関する個々の動作実現, それらを統合した運転実験について述べていく.

  本稿の構成は以下のようになっている.
  \secref{sec:hardware}では, 開発されたMusashiのハードウェア特徴とその自動運転動作への利用について述べる.
  \secref{sec:software}では, 学習型ソフトウェアの概要と, その自動運転動作への適用について述べる.
  \secref{sec:experiments}では, これらを統合し, 認識を含むペダル操作・ハンドル操作実験について述べる.
  \secref{sec:discussion}では, それぞれの実験における課題と今後の方針について議論し, 最後に結論を述べる.
}%

\begin{figure*}[t]
  \centering
  \includegraphics[width=2.0\columnwidth]{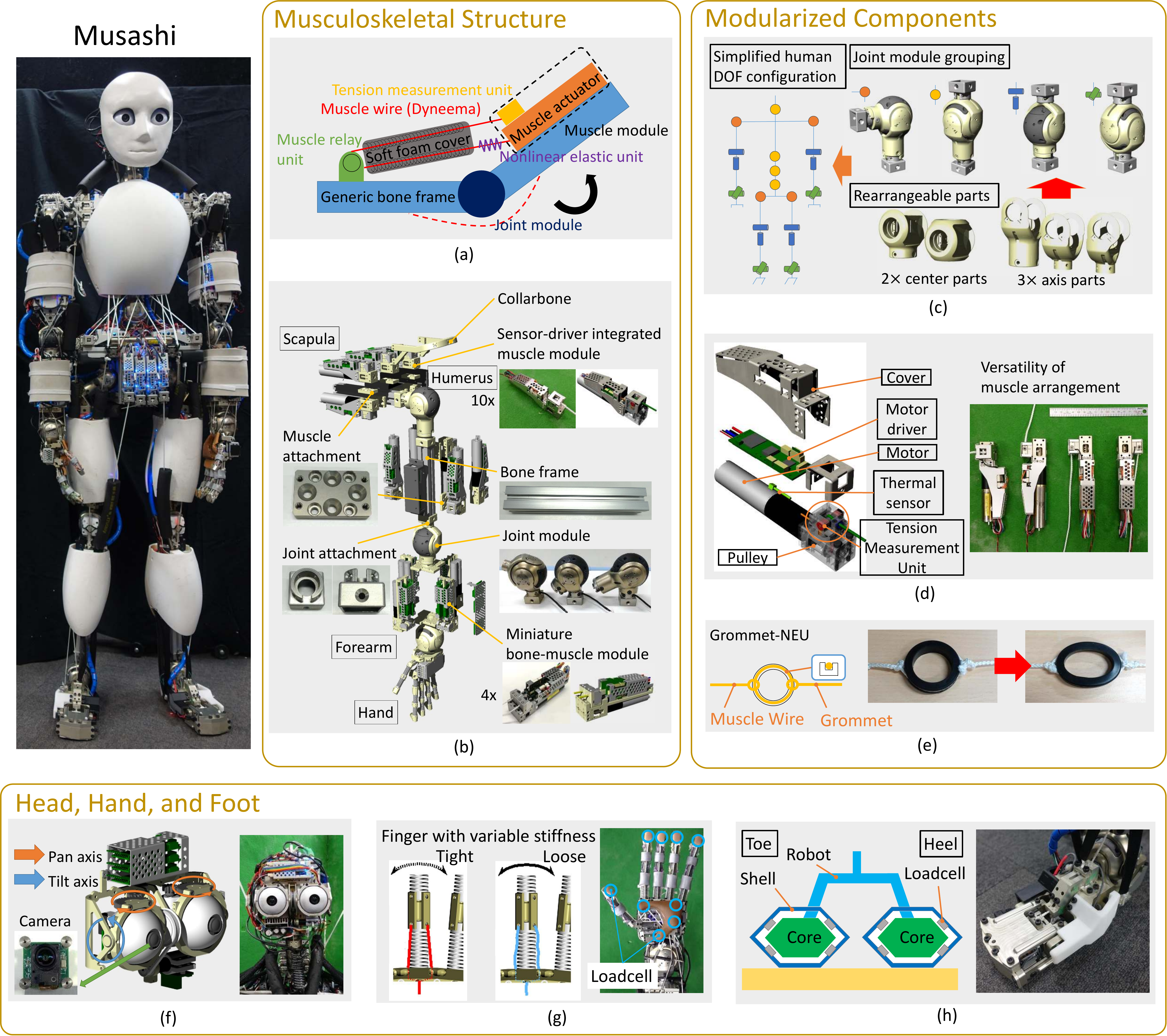}
  \caption{The overview of the developed musculoskeletal humanoid Musashi. (a) shows the basic musculoskeletal structure. (b) shows the modularized structure of the left arm of Musashi combining various modules, (c) shows the rearrangeable concept of joint modules. (d) shows the detailed structure of muscle module (left figure) and the versatility of muscle arrangement (right figure). (e) shows the nonlinear elastic unit. (f) shows the movable eye unit. (g) shows the flexible musculoskeletal hand with machined springs. (h) shows the foot with six-axis core-shell structural force sensors.}
  \label{figure:hardware-overview}
  \vspace{-3.0ex}
\end{figure*}

\section{Hardware of the Musculoskeletal Humanoid} \label{sec:hardware}
\subsection{Design Process} \label{subsec:hardware-process}
\switchlanguage%
{%
  The important points for a robot to drive a car made for humans are considered to be (1) body proportion, (2) body flexibility, and (3) redundant sensors and learning system using them.
  In this paper, (1) we first determine the arrangement of joints and the length of each bone to fit the proportion and joint structure of humans, and attach muscle modules to the bones.
  In addition, (2) a nonlinear elastic unit is attached to the end of each muscle, and the robot can fit into the complex environment with the unit and the original flexibility of muscle wire.
  Finally, (3) we design a structure including not only the muscle module that can measure muscle length, muscle tension, and muscle temperature, but also a flexible hand with contact sensors, movable eyes with high resolution cameras, and loot with 6-axis force sensors that can measure force around their entire surface, as redundant sensors.
  In the next section, we will describe in detail the design of muscle modules and nonlinear elastic units that form (2), and joint modules, eyes, hands and foot that form (3).
}%
{%
  人間が乗るような車を運転するロボットにおいて重要なのは, 人間のような(1)身体プロポーション・(2)身体柔軟性・(3)冗長なセンサとそれを用いた学習であると考えられる.
  本ロボットはまず(1)人体のプロポーション・関節構造に合うように関節の配置とそれぞれの骨の長さを決め, これに対して筋モジュールを貼り付けていく方式を取る.
  また, (2)非線形弾性を筋末端に取り付け, 筋ワイヤ本来の柔軟性と合わせて環境に馴染む柔軟身体の設計を行っている.
  最後に, (3)冗長なセンサとして筋モジュールに筋長・筋張力・筋温度センサを入れるだけでなく, 目には可動眼球・手には接触センサを含む柔軟ハンド・足には周囲全体の力を測定可能な6軸力センサが配置された構造を設計した.
  次章では主に(2)を形づくる筋モジュールと非線形弾性要素, (3)を形作る関節モジュール, 目・手・足の設計について詳細を述べる.
}%

\subsection{Hardware Details} \label{subsec:hardware-overview}
\switchlanguage%
{%
  We show the hardware overview of the developed musculoskeletal humanoid Musashi \cite{kawaharazuka2019musashi} in \figref{figure:hardware-overview}.
  Currently, Musashi has 74 muscles and 39 joints excluding the hand.

  (a) shows the basic musculoskeletal structure of Musashi.
  The joint module connects the generic bone frames, and muscle modules are attached to the frame.
  The abrasion-resistant synthetic fiber Dyneema comes out as a muscle from the tension measurement unit of the muscle module and is folded back by the muscle relay unit.
  The nonlinear elastic unit is attached to the end of the muscle, and a soft foam and spring covers the muscle.
  Muscles are antagonistically arranged around the joint.
  We describe muscles mainly conducting the current task as agonist muscles and the others as antagonist muscles.

  (b) is the structure of the left arm of Musashi constructed by the modules of (c) -- (e) explained subsequently.
  The body structure is easily constructed by the modules, muscle attachments connecting bones and muscles, and joint attachments connecting bones and joints.

  (c) shows the detailed structure of the joint module.
  We can express all the basic human joints by rearranging two center parts and three axis parts.
  Potentiometer, IMU, and a circuit integrating sensor data are packaged in the joint module.
  We can change the body structure easily by transforming this joint module.

  (d) is the detailed structure of the muscle module.
  We use two kinds of muscle modules depending on the body part.
  One of them is a sensor-driver integrated muscle module \cite{asano2015sensordriver}, which can drastically improve reliability and maintainability by packaging a motor, motor driver, thermal sensor, tension measurement unit, etc. into one module.
  The muscle is actuated by winding a muscle wire with a pulley.
  The other is a miniature bone-muscle module \cite{kawaharazuka2017forearm}.
  Although the basic concept is the same with \cite{asano2015sensordriver}, the module is used not only as a muscle but also as a bone frame, and dissipates heat to metal by packaging two smaller actuators into one module and filling the space between the two actuators by metal.
  We can get muscle tension, length, and temperature as sensor data.
  Also, as shown in the right figure of (d), we can realize various muscle routes by rearranging the tension measurement unit.

  (e) shows the nonlinear elastic unit (NEU).
  The previous NEUs are constructed by metal and springs, and the structure is not appropriate for environmental contact.
  Therefore, we realize the nonlinear elasticity using the compression of rubber, by covering the grommet rubber with Dyneema.
  Because the NEU is constructed with only rubber and Dyneema, the structure itself is flexible, and it is suitable for environmental contact.

  (f) is the head of Musashi with the movable eye unit \cite{makabe2018eyeunit}.
  The unit includes three joints: a pan joint in each eye and a tilt joint.
  We use DFK-AFUJ003 (ImagingSource, Inc.) as the eyes, and we can change the image resolution, focus, exposure, etc.
  Although we considered using Lidars/Radars, we decided consistently to mimic the structure of human beings in detail.

  (g) is the hand of Musashi with machined springs \cite{makino2018hand}.
  The hand does not break even if hit by a hammer because of the high flexibility.
  Muscles are antagonistically arranged at proximal phalanges of each finger, and the finger stiffness can be changed by pulling the muscles and compressing the springs.
  Nine loadcells are arranged at each fingertip and the palm of each hand, and contact force can be measured from them.

  (h) is the foot of Musashi with six-axis core-shell force sensors \cite{shinjo2019foot}.
  The core-shell structure is arranged at the toe and heel, and multiple loadcells are arranged between the core and shell.
  This structure can measure all the force applied to the entire shell surface, and so the force to the instep of the foot can be measured.
}%
{%
  本研究で開発された筋骨格ヒューマノイドMusashi \cite{kawaharazuka2019musashi}のハードウェアの概要を\figref{figure:hardware-overview}に示す.

  \figref{figure:hardware-overview}の真ん中に示すのがMusashiの全体像である.
  現在全身で74本の筋肉を持ち, 関節は手を除いて39自由度である.
  Musashiはこれまでの筋骨格ヒューマノイド\cite{nakanishi2013design, asano2016kengoro}とは異なり, モジュール性を高めることで容易な構成・再構成を可能とし, 冗長なセンサや柔軟な身体により学習的に環境接触を伴う動作を行うことを目的としている.

  (a)はMusashiの筋骨格構造を表す.
  骨格となる汎用アルミフレームを関節モジュールが繋ぎ, 骨格には筋モジュールがアタッチメントを介して取り付けられている.
  筋モジュールの筋張力測定ユニットから摩擦に強い合成繊維であるDyneemaが伸び, 筋経由点ユニットを介して折り返されている.
  筋末端には非線形弾性ユニットがつき, それら筋の周りを発泡性のソフトカバーが覆っている.
  筋は関節の周りに拮抗に配置され, 現在のタスクを担う方の筋を主動筋, それ以外を拮抗筋と呼ぶ.

  (b)のJoint Moduleは2つのcenter parts, 3つのaxis partsを組み替えることで, 人間の基本的な関節全てを表現することができる構成とした.
  Joint Moduleの中にはそれぞれの軸にポテンショメータと, IMUと, それらを統合し上位に送信するPotentiometer Control Boardが内包されている.
  ケーブルは全てJoint Moduleの中を通り, 通信用のUSBケーブルが一本外に出ているのみである.
  このJoint Moduleを組み替えることで容易に身体構造を変更することが可能である.

  (c)の筋モジュールは部位によって2種類の筋モジュールを使い分けている.
  一つはSensor-driver integrated muscle module \cite{asano2015sensordriver}で, これはモータ・モータドライバ・温度センサ・筋張力測定ユニット等の要素を一つのパッケージに収めることで信頼性・メンテナンス性を向上させたものである.
  モータの軸についたプーリが筋ワイヤを巻き取ることで筋を駆動する.
  もう一つはMiniature bone-muscle module \cite{kawaharazuka2017forearm}で, 基本的なコンセプトはSensor-driver integrated muscle moduleと同様であるが, 一つのパッケージにより小さな筋を2つ備え, その隙間を金属で満たすことで, 筋としてだけでなく骨格としても使用できること, そして金属に熱を逃がすことで放熱性を高めたものである.
  手の指や首の筋を駆動するような比較的弱い筋張力で良い部位は後者を, それ以外は前者を採用している.
  センサとして筋張力・筋長・筋温度センサが得られる.
  また, (c)の右下図のように, 筋張力測定ユニットの取り付け位置を変更することで, 様々な筋経路を実現することができる.

  (d)はNonlinear Elastic Unit (NEU)である.
  これまでの非線形弾性要素はほとんどが金属で構成される硬いものであったが, 環境接触を伴う動作にはそのような構成は不適切である.
  本研究では, グロメット構造を持つゴムの周りをDyneemaで囲み, ゴムの圧縮を利用することで非線形弾性を実現している.
  ゴムとDyneemaのみの使用のため, それ自体が柔軟であり, 柔軟な環境接触に適する.

  (e)はこれら(a) -- (d)までの要素により構成されたMusashiの左腕の構造である.
  モジュール化された数種類の要素と, 骨格と筋を接続するmuscle attachment, 骨格と関節を接続するjoint attachmentのみで, 簡易に身体を構成できる.

  (f)のHead \cite{makabe2018eyeunit}は可動眼球ユニットを有しており, それぞれの目のpan, そしてtiltの3自由度を有している.
  この3自由度はMusashiにおける唯一の軸駆動である.
  カメラは高性能なDFK-AFUJ003 (ImagingSource, Inc.)を使用しており, 解像度の変更, 焦点の変更, exposureの変更等がsoftware側から可能となっている.

  (g)のHand \cite{makino2018hand}における指は切削ばねをつなげて構成されており, ハンマー等で叩かれても壊れないような柔軟性を持つ.
  指のProximal phalangesにはそれぞれ拮抗に筋が配置されており, これを引っ張ることで切削ばねが圧縮され, 指の剛性が変化するような構造を持つ.
  指先・手のひらには片手で合わせて9個のロードセルが配置されており, 接触時の力を測定することが可能である.

  (h)のFoot \cite{shinjo2019foot}は踵とつま先それぞれにCore-shell構造を持ち, CoreとShellの間にロードセルが複数配置されている.
  よって, ロードセルの情報から踵とつま先それぞれに関して6軸の力覚情報を取得することができる.
  また, Shell上にかかる力全てを測定することができるため, 足の甲にかかった力も測定することが可能である.
}%

\begin{figure}[!ht]
  \centering
  \includegraphics[width=1.0\columnwidth]{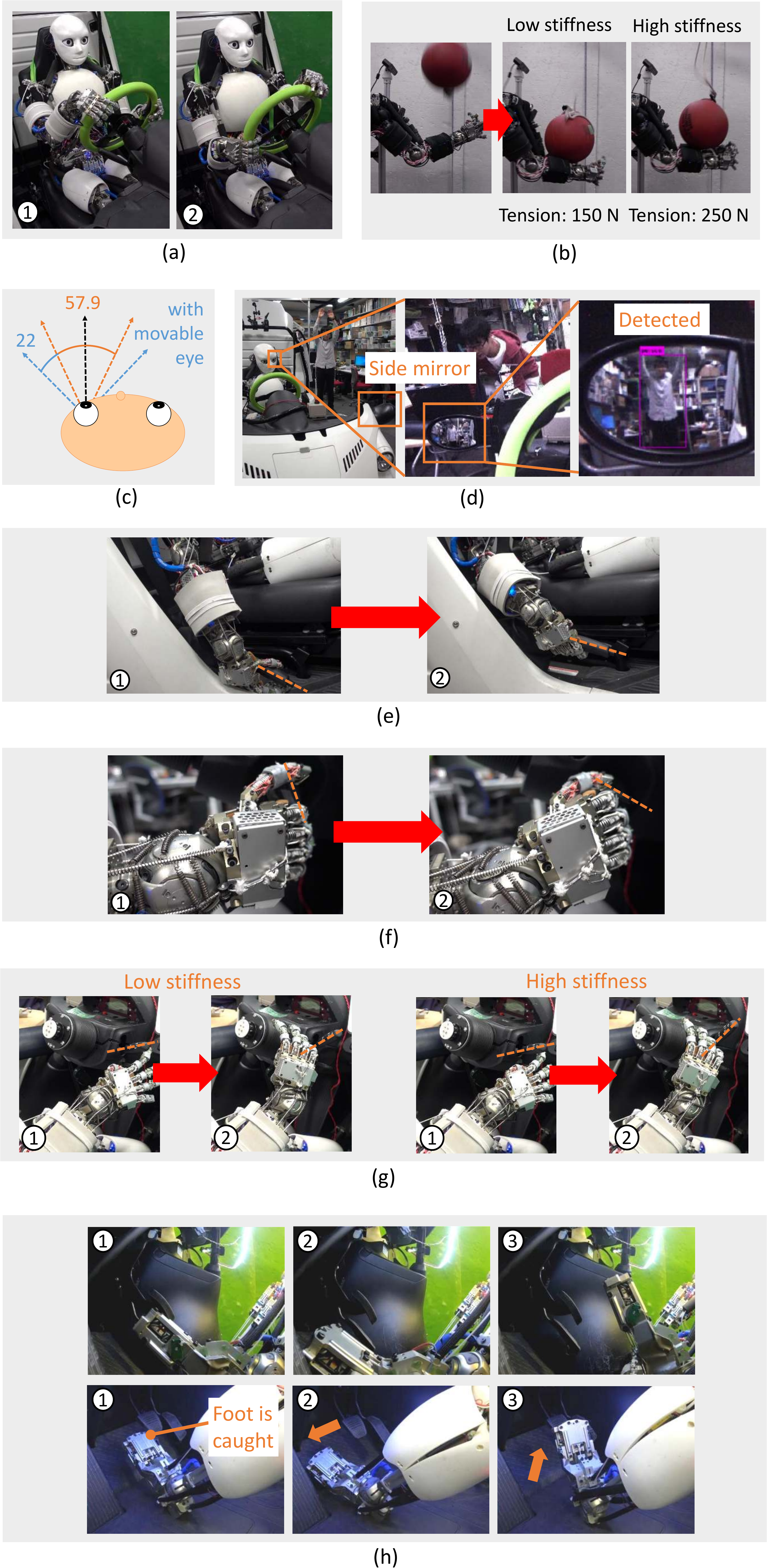}
  \caption{The realization of respective components of autonomous driving using the characteristics of the developed hardware. (a) shows a steering wheel operation with both arms. (b) shows a variable stiffness control using the nonlinear elastic units. (c) shows the field of view of the movable eye unit. (d) shows the human detection experiment in the side mirror using the high resolution camera. (e) shows an experiment pulling a handbrake. (f) shows an experiment rotating a key. (g) shows an experiment operating a blinker lever by changing the stiffness of fingers. (h) shows a recovering experiment from slipping during brake pedal operation using the developed foot.
  }
  \label{figure:hardware-driving}
  \vspace{-3.0ex}
\end{figure}

\begin{figure*}[t]
  \centering
  \includegraphics[width=2.0\columnwidth]{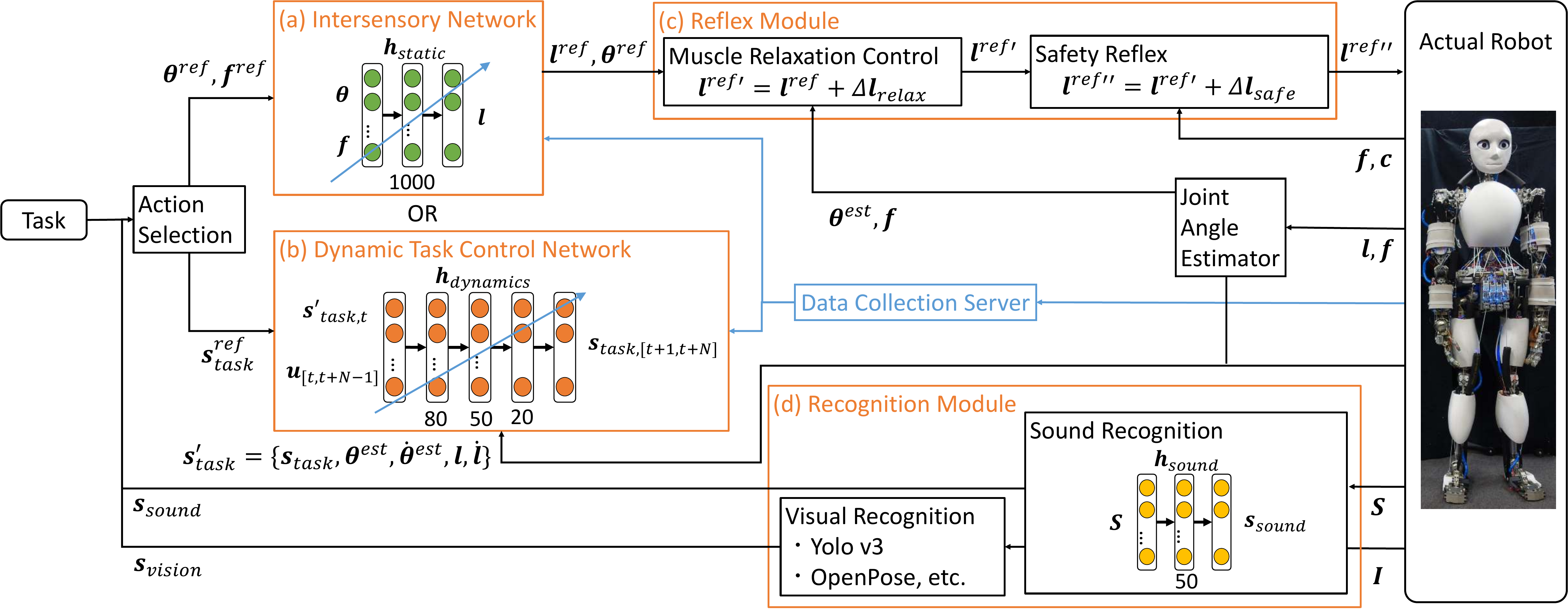}
  \caption{The overview of the developed software with four modules: intersensory network module (static module), dynamic task control network module (dynamic module), reflex module, and recognition module.}
  \label{figure:software-overview}
  \vspace{-3.0ex}
\end{figure*}


\subsection{Application of the Hardware to Autonomous Driving} \label{subsec:hardware-driving}
\switchlanguage%
{%
  We explain the use of the developed nonlinear elastic unit, head, hand, and foot of Musashi in terms of autonomous driving, in \figref{figure:hardware-driving} (presented in multimedia material).

  By using nonlinear elastic units, the robot can not only make the body flexible but can also change the stiffness freely.
  As shown in (a), the flexible body structure makes it possible to operate a steering wheel with both arms, which could not be seen in DRC.
  Also, as shown in (b), the robot can change impact response by changing the body stiffness.
  When dropping a 5 kg ball to the robot from 1 m height, the maximum muscle tension is 150 N with low stiffness, and 250 N with high stiffness.
  This resembles the increase of human arm stiffness during a car crash, and the use of Musashi as a crash test dummy is anticipated.

  By using the head with the movable eye unit, the robot can recognize objects with a wide field of view as shown in (c).
  Also, the robot can recognize a human in a side mirror using the high resolution camera, as shown in (d).

  By using the flexible hand with machined springs, the robot can adapt the grasp shape to various situations inside the car, such as pulling a handbrake (e) and rotating a key (f).
  Also, by using the variable stiffness mechanism of fingers, as shown in (g), the robot can operate a blinker lever correctly with high stiffness.

  By using the foot which can measure the force applied to its entire surface, the robot can recover the foot position by sensing the force to its instep when slipping during a brake pedal operation, as shown in (h).
}%
{%
  Musashiの非線形弾性要素, Head, Hand, Footに関して, 自動運転という観点に着目した利用方法を\figref{figure:hardware-driving}に述べる (presented in multimedia material).

  非線形弾性要素を用いることで, 身体を柔軟にするだけでなく, その柔軟さを自在に変更することができるようになる.
  (a)に示すように, 柔軟な身体構造によって, DARPAでは見ることのなかった両手によるハンドル操作が可能である.
  また, 身体の剛性を変化可能なため, (b)のように, 衝撃に対する応答を自在に変化させることができる.
  手先剛性を小さく設定したとき, 大きく設定したときに1m上から5 kgのボールを落とすと, 前者の最大筋張力は150 N, 後者は250 Nとなった.
  これは車衝突時に人間がグッと力を入れて剛性を高める動作にも似ており, 人体シミュレータとしての利用が期待される.

  可動眼球ユニットを有するHeadを用いることで, (c)のような広い視野範囲による認識行動, また, (d)のような高解像度なカメラを合わせてサイドミラー等に映る車や人の認識も行うことができる.

  切削ばねを用いた柔軟なHandを用いることで, ハンドルだけでなく, (e)のサイドブレーキや(f)の車のキー回し, (g)のウィンカーを上げる動作等様々な形の車内環境に適応することが可能である.
  指の剛性可変機構を用いることで, (g)のように, 柔軟なままでは上げられない車のウィンカーを, 剛性を高くすることで正しく上げることが出来るようになる.
  柔軟性を変化させることで, 様々な動作が可能になるのである.

  (h)では, 面全体の力を測定することができるFootを用いることで, 例えば足がスリップしてブレーキペダルの下に入ってしまったとしても, 足裏の甲にかかる力を感じて元に戻すことができる.
}%

\section{Software of the Musculoskeletal Humanoid} \label{sec:software}
\subsection{Design Process} \label{subsec:software-process}
\switchlanguage%
{%
  In order to handle the flexible body and redundant sensors, we consider that (1) a learning-based motion generation, (2) a learning-based recognition, and (3) a fast reflex control are necessary.
  (1) can be divided into two types of motion generators, for static behavior and dynamic behavior.
  In the former, the dimension of state space is relatively small and online learning is effective.
  In the latter, the dimension is relatively large and only offline learning can be used currently.
  In addition, we also develop (2) learning-based recognition methods using eyes and ear sensors, because visual and sound recognition is significant in performing the driving task.
  Since it is difficult to operate learning-based controls at high frequency, a safety mechanism should be implemented in a lower layer as (3) reflex control.
  In this study, we use a simple conditional branching based on the recognition results for motion planning.
}%
{%
  これまで説明した柔軟な身体と冗長なセンサを扱うためには, (1)学習型の動作生成手法, (2)学習型の認識手法, (3)それらを補う速い周期における反射型制御が必要であると考えた.
  (1)は静的な動作を扱う動作生成手法・動的な動作を扱う動作生成手法に分けることができる.
  前者は状態数が少ないため広く学習を行うことができ, 後者は状態数が増えるため複雑な学習は難しいが動的な要素を扱うことが可能である.
  また, タスクを行う上で認識が必要であり, (2)目や耳のセンサを用いた学習型の認識手法も開発する.
  しかし, 学習型の制御は速い周期で運用することが難しいため, 安全機構等はより低レイヤに(3)反射型制御として実装する.
  これらは基本的なサイクルである認識・計画・動作のうちの認識と動作について扱っており, 本研究では計画については認識結果に基づいた簡易な条件分岐を用いることとした.
}%

\subsection{Software Details} \label{subsec:software-overview}
\switchlanguage%
{%
  We show the overview of the learning-based software system in \figref{figure:software-overview}.
  We describe the current muscle length as $\bm{l}$, the current muscle tension as $\bm{f}$, the current joint angle as $\bm{\theta}$, the current muscle temperature as $\bm{c}$, the muscle Jacobian as $G$, the current image as $\bm{I}$, the current sound as $\bm{S}$, the task state as $\bm{s}_{task}$, the current state obtained from vision as $\bm{s}_{vision}$, and the current state obtained from sound as $\bm{s}_{sound}$.
  Also, $\{\bm{l}, \bm{\theta}, \bm{f}, \bm{s}_{task}\}^{ref}$ is the control command of each state, and $\bm{\theta}^{est}$ is the estimated joint angle.

  The basic components of this system are (a) intersensory network module (static module), (b) dynamic task control network module (dynamic module), (c) reflex module, and (d) recognition module.

  The static module (a) acquires the function $\bm{h}_{static}$ below \cite{kawaharazuka2019longtime}.
  \begin{align}
    \bm{l}=\bm{h}_{static}(\bm{\theta}, \bm{f}) \label{eq:static-network}
  \end{align}
  This represents the static relationship among $\bm{\theta}$, $\bm{l}$, and $\bm{f}$.
  When $\bm{\theta}^{ref}$ and $\bm{f}^{ref}$ are given depending on the task, by inputting them into \equref{eq:static-network}, $\bm{l}^{ref}$ to send to the robot is calculated.
  This function is  expressed by a neural network, and is initialized by man-made data.
  Then, we update the network online by using the actual robot sensor information $(\bm{\theta}, \bm{f}, \bm{l})$.
  At every movement, the network is updated, and the robot becomes able to realize $\bm{\theta}^{ref}$, $\bm{f}^{ref}$ accurately.
  In this study, $\bm{\theta}$ can be obtained from the joint module, but the ordinary musculoskeletal humanoid does not have joint angle sensors due to complex joints such as the ball or scapula joints.
  In that case, $\bm{\theta}$ must be obtained from a motion capture or vision sensor.
  Also, by using this network $\bm{h}_{static}$, not only control but also estimation of joint angles $\bm{\theta}^{est}$ are enabled by using Extended Kalman Filter (EKF) and the change in muscle length and tension.

  The dynamic module (b) acquires the function $\bm{h}_{dynamic}$ below \cite{kawaharazuka2019pedal},
  \begin{align}
    \bm{s}_{task, [t+1, t+N]} = \bm{h}_{dynamics}(\bm{s}'_{task, t}, \bm{u}_{[t, t+N-1]})\label{eq:dynamic-network}
  \end{align}
  where $\{\bm{s}_{task}, \bm{u}\}_{[t_{1}, t_{2}]}$ is a vector vertically arranging $\{\bm{s}_{task}, \bm{u}\}$ from $t_{1}$ to $t_{2}$ time steps, $\bm{s}'_{task, t}$ is the initial task state ($\{\bm{s}_{task}, \bm{\theta}^{est}, \dot{\bm{\theta}}^{est}, \bm{l}, \dot{\bm{l}}\}$ in this study), $\bm{u}$ is the control command ($\bm{\theta}^{ref}$ or $\bm{l}^{ref}$ in this study), and $N$ is the number of time steps to expand this state equation.
  This function is equivalent to a network representing a dynamic transition of the task state by a control command sequence.
  This network can be trained by gathering data of observed task state transition when sending random control commands.
  When realizing a certain task state $\bm{s}_{task}^{ref}$, we execute the equations below after setting the initial $\bm{u}$,
  \begin{align}
    \bm{s}^{pred}_{task, seq} &= \bm{h}_{dynamic}(\bm{s}'_{task, t}, \bm{u}^{init}_{seq})\\
    L &= \textrm{MSE}(\bm{s}^{pred}_{task, seq}, \bm{s}_{task, seq}^{ref}) + \alpha{E}_{adj}(\bm{u}^{init}_{seq}) \label{eq:dynamic-loss}\\
    \bm{g} &= dL/d\bm{u}^{init}_{seq} \\
    \bm{u}^{init}_{seq} &\gets \bm{u}^{init}_{seq} - \beta\frac{\bm{g}}{|\bm{g}|}
  \end{align}
  where $\bm{u}^{init}_{seq}$ is the sequence of the initial value of control command, $\bm{s}^{pred}_{task, seq}$ is the predicted sequence of $\bm{s}_{task}$ in $[t+1, t+N]$ calculated from $\bm{s}'_{task, t}$ and $\bm{u}^{init}_{seq}$, and $\bm{s}^{ref}_{task, seq}$ is a vector arranging $N$ number of $\bm{s}^{ref}_{task}$.
  Also, $\textrm{MSE}$ is mean squared error, $E_{adj}$ is MSE of the values at adjacent time steps, $\alpha$ is a weight constant, and $\beta$ is an update rate.
  We calculate the loss between the predicted and target $\bm{s}_{task}$, add the loss to smoothen the target control command sequence, and update the control command sequence by backpropagation \cite{rumelhart1986backprop}.
  By repeating this procedure, $\bm{u}^{init}_{seq}$ is updated to accurately realize $\bm{s}^{ref}_{task}$, and the task is executed by sending $\bm{u}^{init}_{t}$ to the actual robot.
  While (a) is the network regarding static motions, (b) is the network regarding dynamic motions, and its learning is difficult due to more variables.
  So, the robot is moved mainly by (a), but (b) is constructed offline and used when conducting tasks in which dynamic and accurate motions are required.

  (c) is the reflex module which is executed at high frequency.
  Muscle Relaxation Control (MRC) \cite{kawaharazuka2019relax} is a control elongating muscles from antagonist muscles to agonist muscles while keeping the current joint angle.
  First, we calculate the necessary muscle tension $\bm{x}$ achieving the necessary joint torque $\bm{\tau}^{nec}$ by solving quadratic programming.
  \begin{align}
    \underset{\bm{x}}{\textrm{minimize}}&\;\;\;\;\;\;\;\bm{x}^{T}W_{1}\bm{x} + (G^{T}\bm{x}+\bm{\tau}^{nec})^{T}W_{2}(G^{T}\bm{x}+\bm{\tau}^{nec})\\
    \textrm{subject to}&\;\;\;\;\;\;\;\;\;\;\;\;\;\;\;\;\;\;\;\;\;\;\;\;\;\;\;\;\;\bm{x} \geq \bm{f}^{min}
  \end{align}
  where $W_{1}, W_{2}$ are weight matrices, and $\bm{f}^{min}$ is the minimum muscle tension.
  We sort muscles by $\bm{x}$ in ascending order, and gradually elongate them in order starting with unnecessary antagonist muscles with smaller tension.
  In detail, we gradually increase the muscle relaxation value $\Delta\bm{l}_{relax}$, and add it to the muscle length to send to the actual robot.
  When the muscle tension becomes smaller than $\bm{f}^{min}$, the muscle to elongate is changed to the next muscle, and this control stops when the current joint angle is changed more than a threshold.
  This procedure works in a static state.
  At a moving state, we sort muscles by $\bm{x}$ in descending order, and gradually decrease $\Delta\bm{l}_{relax}$ starting with necessary muscles.
  This reflex can inhibit unnecessary muscle tension of antagonist muscles due to the model error.
  Also, for example, when resting the arms on the table, the body and environment are constrained, and so not only antagonist muscle tension but also agonist muscle tension can be reduced, because the joint angle does not change even when relaxing agonist muscles.
  Thus, the robot becomes able to rest its body and move continuously for a longer time.
  Safety reflex is a simple control that elongates muscle length in order not to break motors due to high muscle tension and temperature.
  We add $\Delta\bm{l}_{safe}$ calculated as below to the target muscle length,
  \begin{align}
    \Delta\bm{l}^{ref}_{safe} =& K_{f}\textrm{max}(\bm{f}-\bm{f}^{lim}, \bm{0})+K_{c}\textrm{max}(\bm{c}-\bm{c}^{lim}, \bm{0})\\
    \Delta\bm{l}_{safe} \gets& \Delta\bm{l}_{safe} + \textrm{max}(\Delta\bm{l}^{min}, \textrm{min}(\Delta\bm{l}^{max}, \Delta\bm{l}^{ref}_{safe}-\Delta\bm{l}_{safe}))
  \end{align}
  where $\{\bm{f}, \bm{c}\}^{lim}$ is a threshold of muscle tension or temperature to begin the reflex, $K_{\{f, c\}}$ is a gain, $\Delta\bm{l}^{ref}_{safe}$ is an ideal elongation value, and $\Delta\bm{l}^{\{min, max\}}$ is a minimum or maximum change of $\Delta\bm{l}_{safe}$ at one time step.
  By elongating the muscle length considering muscle tension and temperature, we can inhibit the burnout of motors.

  (d) is the recognition module, and it is divided into object and sound recognition.
  In this study, we use Yolo v3 \cite{redmon2018yolov3} for object recognition.
  The labels mainly used in this study is ``car,'' ``person,'' and ``traffic light.''
  Also, regarding sound recognition, the sound is converted into a mel spectrum, and the network $\bm{h}_{sound}$ is trained to output sound class from the spectrum.
  In this study, this module recognizes only noise or car horn.
}%
{%
  Musashiの学習型ソフトウェアシステムの概要を\figref{figure:software-overview}に示す.
  ここで, $\bm{l}$は現在筋長, $\bm{f}$は現在筋張力, $\bm{\theta}$は現在関節角度, $\bm{c}$は現在筋温度, $G$は筋長ヤコビアン, $\bm{I}$は現在画像, $\bm{S}$は現在音声, $\bm{s}_{task}$は設定したタスクの状態を表す変数, $\bm{s}_{vision}$は視覚から得られた現在状態, $\bm{s}_{sound}$は音声から得られた現在状態を表す.
  また, $\{\bm{l}, \bm{\theta}, \bm{f}, \bm{s}_{task}\}^{ref}$はそれぞれの状態の制御指令値, $\bm{\theta}^{est}$は関節角度の推定値を表す.

  基本的な要素は4つであり, それぞれ, intersensory network module (static module), dynamic task control network module (dynamic module), reflex module, recognition moduleである.
  それらの概要を\figref{figure:software-respective}に示す.
  特に(a), (b)のモジュールは, 筋骨格ヒューマノイドの柔軟な身体構造はモデル化が難しく, 身体の動かし方を実機センサデータに基づいて獲得していく必要があることに起因する.
  (c)は反射型の安全管理制御, (d)は目・耳による認識に関するモジュールになっている.

  (a)のintersensory network module (static module) \cite{kawaharazuka2019longtime}は以下の関数$\bm{h}_{static}$を学習していくことに相当する.
  \begin{align}
    \bm{l}=\bm{h}_{static}(\bm{\theta}, \bm{f}) \label{eq:static-network}
  \end{align}
  これは$\bm{\theta}$, $\bm{l}$, $\bm{f}$の静的な関係を表す.
  Taskを与えると, それに応じた$\bm{\theta}^{ref}$や$\bm{f}^{ref}$が決まる場合, それを\equref{eq:static-network}に代入することで, ロボットに送るべき$\bm{l}^{ref}$を算出することができる.
  このnetwork(本研究ではneural networkで表現する)は, 始めは人間が作成する.
  CAD上で筋の起始点・中継点・終止点を直線で結んだものを筋経路とし, 筋張力による非線形弾性要素やDyneemaの伸びをモデル化することでデータセットを作成し, このnetworkを学習させる.
  しかし, CADから得たモデルには大きな誤差が存在するため, これを実機センサデータからより正しく更新していく必要がある.
  そこで, 実機から$(\bm{\theta}, \bm{f}, \bm{l})$の情報を取得し, これを用いて$\bm{h}_{static}$をオンラインで更新していく.
  すると, 動作するごとにnetworkが更新されていき, より正確に$\bm{\theta}^{ref}$, $\bm{f}^{ref}$を実現できるようになるのである.
  本研究では$\bm{\theta}$を関節モジュールから得ることができるが, 通常の筋骨格ヒューマノイドは球関節等の影響から関節角度センサを備えていない.
  その場合は, motion captureや視覚センサ情報から$\bm{\theta}$を得る必要がある.
  また, このネットワーク$\bm{h}_{static}$を用いることで, 制御だけでなく, 筋長・筋張力変化から拡張カルマンフィルタ(EKF)を用いて関節角度$\bm{\theta}^{est}$の推定を行うことができる.

  (b)のdynamic task control network module (dynamic module) \cite{kawaharazuka2019pedal}は以下の関数$\bm{h}_{dynamic}$を学習していくことに相当する.
  \begin{align}
    \bm{s}_{task, [t+1, t+N]} = \bm{h}_{dynamics}(\bm{s}'_{task, t}, \bm{u}_{[t, t+N-1]})\label{eq:dynamic-network}
  \end{align}
  ここで, $\bm{a}_{[A, B]}$は$[A, B]$の間におけるベクトル$\bm{a}$を縦に並べたベクトル, $\bm{s}'_{task, t}$は初期状態(本研究では$\{\bm{s}_{task}, \bm{\theta}^{est}, \dot{\bm{\theta}}^{est}, \bm{l}, \dot{\bm{l}}\}$とする), $\bm{u}$はロボットの制御入力(本研究では$\bm{\theta}^{ref}$または$\bm{l}^{ref}$), $N$は時系列に展開するタイムステップ数を表す.
  これは, タスク状態の制御入力による動的な遷移を表すネットワークに相当する.
  このnetworkを実機のデータから学習する.
  例えばランダムな制御入力を加えたときのタスク状態変化を観察しデータセットを作成することで, このネットワークを構築することができる.
  あるタスク状態$\bm{s}_{task}^{ref}$を実現したいときは, 適当な$\bm{u}$の初期値を決定し, 以下の式を実行する.
  \begin{align}
    \bm{s}^{pred}_{task, seq} &= \bm{h}_{dynamic}(\bm{s}'_{task, t}, \bm{u}^{init}_{seq})\\
    L &= \textrm{MSE}(\bm{s}^{pred}_{task, seq}, \bm{s}_{task, seq}^{ref}) + \alpha{E}_{adj}(\bm{u}^{init}_{seq}) \label{eq:dynamic-loss}\\
    \bm{g} &= dL/d\bm{u}^{init}_{seq} \\
    \bm{u}^{init}_{seq} &\gets \bm{u}^{init}_{seq} - \beta\frac{\bm{g}}{|\bm{g}|}
  \end{align}
  ここで, $\bm{u}^{init}_{seq}$は制御入力の初期値$\bm{u}^{init}$を並べた$\bm{u}^{init}_{[t, t+N-1]}$, $\bm{s}^{pred}_{task, seq}$は$\bm{s}'_{task, t}$と$\bm{u}^{init}_{seq}$から予測された$[t+1, t+N]$の$\bm{s}_{task}$の時系列, $\bm{s}^{ref}_{task, seq}$はN個$\bm{s}^{ref}_{task}$を並べたものを表す.
  また, $\textrm{MSE}$はMean Squared Errorを, $E_{adj}$は隣り合う時系列間の値の二乗誤差平均, $\alpha$は係数を表す.
  予測された$\bm{s}_{task}$の時系列と指令値の間の誤差を取り, また, 隣り合う制御入力が近くなるように, つまり滑らかに制御入力を変えていくように誤差を定義し, これを制御入力に対して誤差逆伝播\cite{rumelhart1986backprop}していく.
  これを繰り返すことで, $\bm{u}^{init}_{seq}$はより正確に$\bm{s}^{ref}_{task}$を実現していくものへと変化していき, $\bm{u}^{init}_{t}$を実機に送ることで, タスクを実行する.
  (a)は静的な動作に関するネットワークなのに対して, (b)は動的な動作に関するものであり, より変数が多くなるため学習が難しい.
  そのため, 基本的には(a)を用いて動作するが, より動的で正確な動作が要求されるタスクにおいては, (b)をオフラインで構築し実行する.

  (c)のreflex moduleは, muscle relaxation control, safety reflexの2つからなる, 速い周期で実行される制御群である.
  muscle relaxation control \cite{kawaharazuka2019relax}は, 現在の関節角度を保持したまま拮抗筋から主動筋までを順に緩ませていく制御である.
  まず, タスクに必要な関節トルク$\bm{\tau}^{nec}$から以下のQuadratic Programmingを解くことで必要な筋張力$\bm{x}$を導出する.
  \begin{align}
    \underset{\bm{x}}{\textrm{minimize}}&\;\;\;\;\;\;\;\bm{x}^{T}W_{1}\bm{x} + (G^{T}\bm{x}+\bm{\tau}^{nec})^{T}W_{2}(G^{T}\bm{x}+\bm{\tau}^{nec})\\
    \textrm{subject to}&\;\;\;\;\;\;\;\;\;\;\;\;\;\;\;\;\;\;\;\;\;\;\;\;\;\;\;\;\;\bm{x} \geq \bm{f}^{min}
  \end{align}
  ここで, $W_{1}, W_{2}$は重み行列であり, $\bm{f}^{min}$は最小の筋張力を表す.
  これを昇順にソートし, より小さい筋張力, つまり必要のない拮抗筋から順に少しずつ緩ませていく.
  具体的には, 弛緩度合い$\Delta\bm{l}_{relax}$を増やしていき, 最終的に実機に送る筋長にそれを加算する.
  筋張力が$\bm{f}^{min}$より小さくなれば, 緩める筋を次の筋へと移動していき, 関節角度がある一定以上変化してしまったらこの制御を止める.
  本制御はロボットの静止時にのみ働き, 逆に動作時には筋張力を降順に並べ, 必要のある筋から$\Delta\bm{l}_{relax}$を減らしていく.
  これにより, モデル誤差による無駄な拮抗筋にかかる筋張力を抑えることができる.
  また, 例えば手を机の上に置くような動作では, 身体と環境が拘束されるため, 拮抗筋だけでなく, 主動筋を緩ませても関節角度は変化しない.
  ゆえに, 拮抗筋だけでなく主動筋の筋張力も削減でき, より身体をしっかりと休め, 長時間の行動が可能となっていく.
  safety reflexは非常に単純で, 高い筋張力・筋温度によりモータが破損しないようにするための制御である.
  以下のように計算された$\Delta\bm{l}_{safe}$を指令筋長へと加算する.
  \begin{align}
    \Delta\bm{l}^{ref}_{safe} =& K_{f}\textrm{max}(\bm{f}-\bm{f}^{lim}, \bm{0})+K_{c}\textrm{max}(\bm{c}-\bm{c}^{lim}, \bm{0})\\
    \Delta\bm{l}_{safe} \gets& \Delta\bm{l}_{safe} + \textrm{max}(\Delta\bm{l}^{min}, \textrm{min}(\Delta\bm{l}^{max}, \Delta\bm{l}^{ref}_{safe}-\Delta\bm{l}_{safe}))
  \end{align}
  ここで, $\{\bm{f}, \bm{c}\}^{lim}$はsafety reflexを駆動させ始める筋張力・筋温度の閾値, $K_{\{f, c\}}$は係数, $\Delta\bm{l}^{ref}_{safe}$は伸張させるべき理想の値, $\Delta\bm{l}^{\{min, max\}}$は1 timestepにおける$\Delta\bm{l}_{safe}$の変化の最小値と最大値を表す.
  このように筋長変化を制限しながら筋温度と筋張力を見て弛緩させることで, 振動を防ぎつつ, モータの破損を抑制することができる.

  (d)のrecognition moduleは物体の認識と, 音声の認識に分けられる.
  本研究では, 物体認識はYolo v3 \cite{redmon2018yolov3}を用いている.
  主に用いるラベルは, car, human, traffic lightの3つである.
  また, 音声認識は音声をメルスペクトラムに変換し, それを入力として音声のクラスを出力するようなネットワーク$\bm{h}_{sound}$を学習させている.
  本研究では主にnoiseとcar hornのみを識別している.
}%

\begin{figure}[!ht]
  \centering
  \includegraphics[width=1.0\columnwidth]{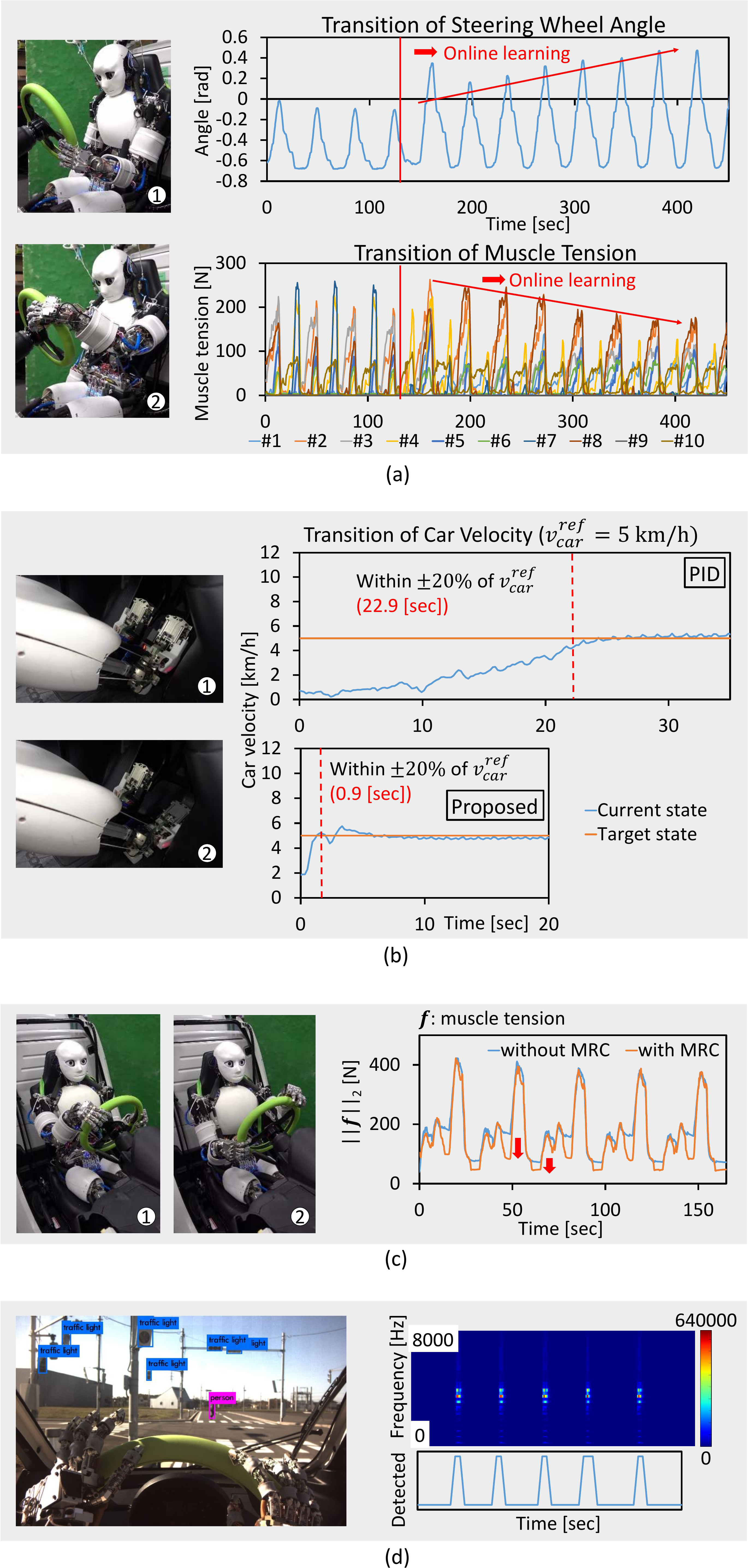}
  \caption{Respective components of autonomous driving using the developed software. (a) shows a steering wheel operation experiment using the online learning of static module. (b) shows a pedal operation experiment using the trained dynamic module. (c) shows a steering wheel operation experiment with and without muscle relaxation control. (d) shows visual recognition of traffic lights and a human, and sound recognition of a car horn.}
  \label{figure:software-driving}
  \vspace{-3.0ex}
\end{figure}

\subsection{Application of the Software to Autonomous Driving} \label{subsec:software-driving}
\switchlanguage%
{%
  We show the use of each software component for autonomous driving in \figref{figure:software-driving} (presented in multimedia material).

  The static module can be applied to various motions, and especially, is useful for steering wheel operation.
  In (a), we show an experiment to operate the steering wheel during online learning of the static module.
  The graphs of (a) show the transition of the steering wheel angle and 10 muscle tensions of the shoulder and elbow of Musashi.
  At every operation, the angle gradually increases, and muscle tension gradually decreases.
  Thus, the relationship among $\bm{f}$, $\bm{\theta}$, and $\bm{l}$ is correctly updated using the actual robot sensor information.

  The dynamic module is useful for dynamic motions.
  The pedal operation is a good example because a fast motion adaptation is necessary.
  In (b), we trained the dynamic module using the data of random pedal operation for one minute and conducted pedal operation with the trained network.
  This experiment was conducted indoors and the rear wheels were on free rollers for safety.
  The car velocity was obtained through CAN-USB of the car.
  We describe the car velocity as $v_{car}$ and the joint angle of right ankle pitch as $\theta_{ankle}$, and set $\theta_{ankle}$ as $\bm{u}$.
  We set $v^{ref}_{car}=5 [km/h]$, and conducted the manually tuned PID control (\textbf{PID}) and the proposed control with the dynamic module (\textbf{Proposed}).
  When comparing \textbf{PID} and \textbf{Proposed}, the error between $v_{car}$ and $v^{ref}_{car}$ falls within $20\%$ starting from 22.9 sec and 0.9 sec, respectively.
  Since the dynamic relationship between the joint angle of ankle-pitch and car velocity is complex, this was the limit of \textbf{PID} by manual tuning and the car velocity vibrated largely when further increasing the gain in our experiment.
  Thus, by acquiring the state equation between task state and control command, fast tracking is enabled.

  The reflex module is useful for motions with long rest time.
  For example, as shown in (c), during the steering wheel operation, the robot does not always operate the wheel.
  In this case, Muscle Relaxation Control (MRC) gradually makes antagonist muscles elongate and makes internal force decrease.
  Also, because the steering wheel and the hands are constrained, the current joint angle does not change largely even when elongating agonist muscles.
  We show the transition of L2 norm of 10 muscle tensions of the elbow and shoulder in the left arm of Musashi $||\bm{f}||_{2}$, with and without MRC.
  The muscle tension is reduced in a static state with MRC compared to without MRC.
  This module can inhibit the increase of muscle temperature, and long-time operation is enabled.

  The recognition module is mainly used as visual recognition of ``person'' and ``traffic light,'' and sound recognition of the car horn.
  The left figure of (d) shows the recognition result at a crossing, and the traffic lights and a human are recognized well.
  Also, the right figure of (d) shows the sound spectrum of a car horn, and the car horn is recognized well.
}%
{%
  それぞれの制御コンポーネントの, 自動運転に着目した利用について\figref{figure:software-driving}に示す (presented in multimedia material).

  static moduleは様々な動作において適用可能であるが, 特に, 環境との接触を要するハンドル操作において有用である.
  (a)ではオンライン学習を実行させると同時にハンドルを回し続ける実験を行った.
  その際のハンドル角度の遷移, Musashiの肩と肘に備わる10本の筋の筋張力の遷移を表している.
  動作を継続するごとに徐々に回転するハンドルの角度が大きく, また, 筋張力が徐々に下がっていることがわかる.
  よって, 動作する中で実機における$\bm{f}, \bm{\theta}, \bm{l}$の関係が正しく学習されていくことがわかる.

  dynamic moduleは動的な動作に有用であるが, その中でもペダル操作は適応的な速度調整が必要であり応用が期待される.
  (b)では1分間程度ランダムにペダルを踏み走行したときのデータを用いてDynamic Task Control Networkを学習させ, それを用いてペダル操作を行う.
  また, 本実験は車の後輪をフリーローラの上に起き, 室内で実験している.
  ここで, $v_{car}$は車速, $\theta_{ankle}$は足首の関節角度を表し, $s_{task}=v_{car}$, $\bm{u}=\theta_{ankle}$とする.
  $v^{ref}_{car}=5 [km/h]$と設定したうえで, 通常のPID制御(\textbf{PID}), Dynamic Task Control Network (\textbf{Proposed})を用いた制御を実行した.
  関節角度$\theta^{ref}_{ankle}$は最終的にstatic moduleによって筋長$\bm{l}^{ref}$に変換されて実機に送られている.
  手動でチューニングされた\textbf{PID}と\textbf{Proposed}を比べると, それ以降$v_{car}$と$v^{ref}_{car}$の誤差が$20\%$以下になるような時間は22.9 secと0.9 secとなった.
  よって, タスク状態と制御入力に関する状態方程式を獲得して最適制御を実行することで, より速い制御指令値への追従が可能となった.

  reflex moduleは静止するような動作が多い場合は非常に有効である.
  例えば(c)のような両手での運転動作ではずっとハンドルを回転させているわけではなく, 真っ直ぐの道ではハンドルをほとんど回さないことが多い.
  その場合, Muscle Relaxation Control (MRC)によって拮抗筋が徐々に緩み, 内力が減っていく.
  また, ハンドルという環境に手をかけているため主動筋を弛緩させても現在関節はほとんど変わらない.
  グラフにはMRCを入れた場合と入れない場合における左腕の肩と肘の10本の筋の筋張力のL2ノルム$||\bm{f}||_{2}$の遷移を示す.
  MRCを入れた場合は入れない場合に比べて静止時で筋張力が削減されていることがわかる.
  これにより, 筋温度の上昇が抑えられ, より長時間の動作が可能となる.

  recognition moduleは本研究では主に交差点における人・信号の認識, car hornの認識に利用する.
  (d)の左図は交差点における認識結果であるが, しっかりと信号や人等を認識できていることがわかる.
  また, (d)の右図では車のクラクションの際の音声スペクトラムとその際の認識結果を示している.
  正しくcar hornを認識できていることがわかる.
}%

\begin{figure}[t]
  \centering
  \includegraphics[width=1.0\columnwidth]{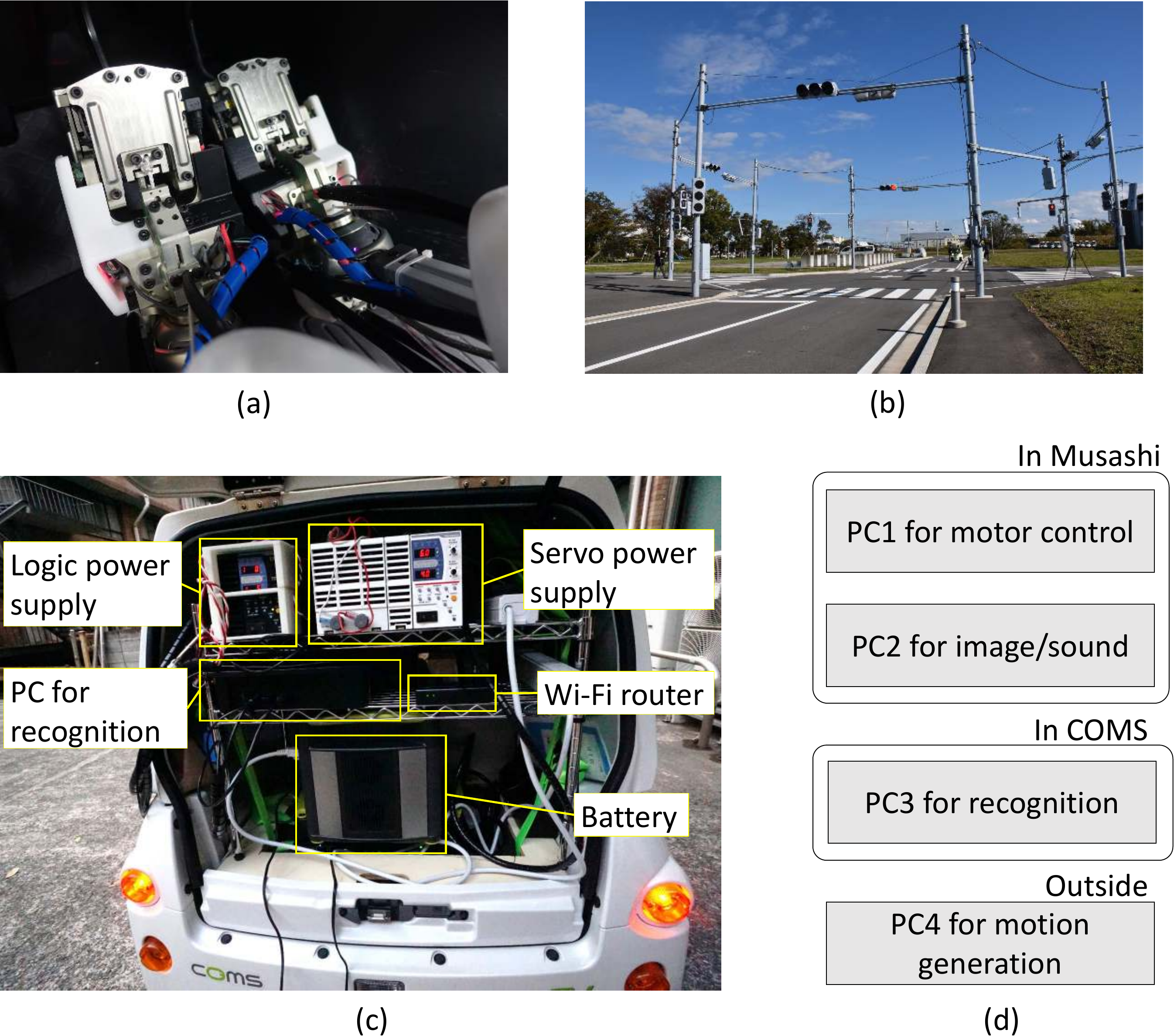}
  \caption{Experimental setup of autonomous driving by the musculoskeletal humanoid Musashi: (a) pedal operation configuration, (b) experimental environment, (c) experimental configuration in COMS, and (d) configuration of PCs.}
  \label{figure:experimental-setup}
  \vspace{-3.0ex}
\end{figure}

\begin{figure*}[t]
  \centering
  \includegraphics[width=2.0\columnwidth]{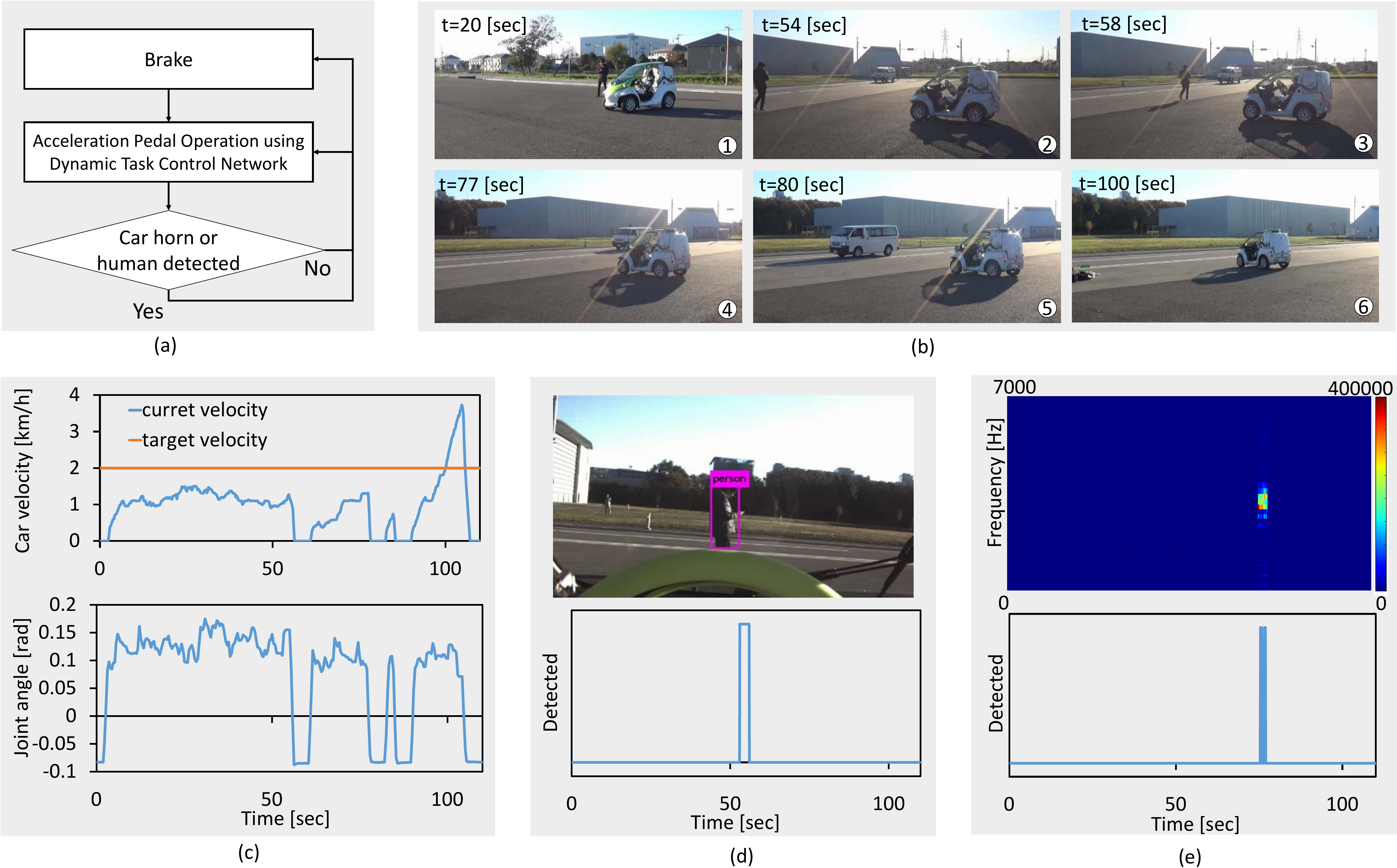}
  \caption{Pedal operation experiment with recognition. (a) shows the experimental motion flow. (b) shows the experimental appearance. (c) shows the current and target car velocity (upper graph) and the joint angle of the right ankle pitch for the acceleration pedal (lower graph). (d) shows the visual recognition result. (e) shows the sound spectrum and its recognition result.}
  \label{figure:pedal-with-recog}
  \vspace{-3.0ex}
\end{figure*}

\section{Experiments} \label{sec:experiments}
\subsection{Experimental Setup} \label{subsec:experimental-setup}
\switchlanguage%
{%
  The car used in this study is a B.COM Delivery of extremely small EV COMS (Chotto Odekake Machimade Suisui) series.
  For safety, its motor torque is limited to 5 Nm, and an emergency stop button is equipped.
  In the pedal operation, as shown in (a) of \figref{figure:experimental-setup}, acceleration and brake pedal is operated by the right and left foot, respectively.
  The experiment is conducted in the Kashiwa Campus at the University of Tokyo, as shown in (b).
  As shown in (c), COMS is equipped with a battery, logic power supply, servo power supply, Wi-Fi router, and PC for recognition module in the trunk.
  COMS is an electric vehicle, and all the electric power for the robot can be obtained from COMS in the future.
  As shown in (d), PCs (Intel NUC, Intel, Inc.) for motor control and image/sound are in the head of Musashi, the recognition is executed on the PC (ZOTAC VR GO, ZOTAC, Inc.) in COMS, and the other processes are executed on PC4 outside.
}%
{%
  本研究のペダル操作実験で用いる自動車は, トヨタ車体製の超小型EV コムス (COMS: Chotto Odekake Machimade Suisui)シリーズのB・COMデリバリーである.
  安全のため, モータトルクはソフトウェア上で5 Nmに制限されており, 非常停止ボタンが備えられている.
  ペダル操作は(a)に示すように右足左足でそれぞれアクセルとブレーキを踏むようになっている.
  実験は(b)の東京大学柏キャンパスの実験施設で行った.
  (c)に示すように, COMSの内部にはバッテリー, ロジックの電源装置, サーボの電源装置, Wi-Fi Router, recognition module用のPCが積まれている.
  COMSは電気自動車であり, 将来的にはロボットの電源は全て車体から得られるようになる可能性がある.
  (d)に示すように, Musashiの頭部にはmotor control用, image/soundの取得用のPCが備えられており, recognitionはCOMS搭載のPC(VRGO)上で行い, それ以外の動作生成関係は外部のPC4で行っている.
}%

\subsection{Pedal Operation with Recognition} \label{subsec:pedal-operation}
\switchlanguage%
{%
  We conducted an experiment integrating the pedal operation and recognition (\figref{figure:pedal-with-recog}, presented in multimedia material).
  We show the experimental motion flow in (a) of \figref{figure:pedal-with-recog}.
  First, the robot operates the acceleration pedal using the trained dynamic module, steps on the brake pedal if it recognizes a human, restarts the acceleration pedal operation, and steps on the brake pedal again if it recognizes a car horn.
  The motion sequence is shown in (b) of \figref{figure:pedal-with-recog}, and we can see that it succeeded.
  (c) shows the transition of car velocity (upper graph) and joint angle of the right ankle pitch (lower graph).
  (d) shows the recognition result of a human.
  The robot recognizes a human when its bounding box is larger than a certain threshold and its center coordinate is around the center of the image.
  (e) shows the recognition result of a car horn.
  When a human (c) or car horn (d) was detected, the brake pedal was stepped on, and we can see that the car velocity became 0.
  The big problem here is the poor tracking of the target car velocity as shown in (c).
  This is because of the difference between the environment that the network is trained in and the experimental environment.
  When $t=[0, 60]$ [sec], the car velocity decreased because the road friction was high, and when $t=[90, 110]$ [sec], the car velocity increased because the road was downhill.
  This was not a problem in the experiment on the free roller, but the difference of the actual environment was difficult to handle.
  Because the learning-based controls depend on the data used for training, we must continue to conduct experiments outside, obtain data, and solve this problem.
}%
{%
  Pedal操作, Recognitionを統合した実験を行った(\figref{figure:pedal-with-recog}, presented in multimedia material).
  動作の遷移図を\figref{figure:pedal-with-recog}の(a)に示す.
  Dynamic Task Control Networkを用いたペダル操作で走行を行い, 人を認識したらブレーキをかけ, また走りだし, クラクションを聞いたらブレーキをかけ, また走りだす.
  それら一連の動作が, \figref{figure:pedal-with-recog}の(b)に示され, 成功していることがわかる.
  (c)はそれぞれ車速(上図)とアクセルペダルを踏み込む右足首ピッチの角度(下図)を示している.
  (d)は人認識の様子とそのときの認識結果を示している.
  人認識は, 人のbounding boxがある閾値よりも大きく, かつ中心にあるときだけ人として認識している.
  また, (e)はクラクションのスペクトルとその認識結果を示しており, 認識が成功していることがわかる.
  (d), (e)におけるそれぞれの検知時に, ブレーキが踏まれ, 車速が0になっていることがわかる.
  ここで大きな問題は, (c)に示したように, 指令車速に対して, 現在車速の追従が悪い点である.
  これは, Dynamic Task Control Networkを学習させた環境と, 実際に走行を行った環境の違いによるものである.
  $t=[0, 60]$ [sec]ではより摩擦が大きく進みにくい路面なため速度が落ち, $t=100$ [sec]では下り坂になっており加速が大きくなってしまっていると考えられる.
  実験室の中でフリーローラの上に車体を載せているときには問題にならなかった, 環境の違いが大きく問題となっている.
  学習型の制御は動作が学習したデータに依存するため, 実験室の中だけでなく, より屋外での実験を重ね, データの取得とこれらの問題を解決できる手法が今後望まれる.
}%

\begin{figure*}[t]
  \centering
  \includegraphics[width=2.0\columnwidth]{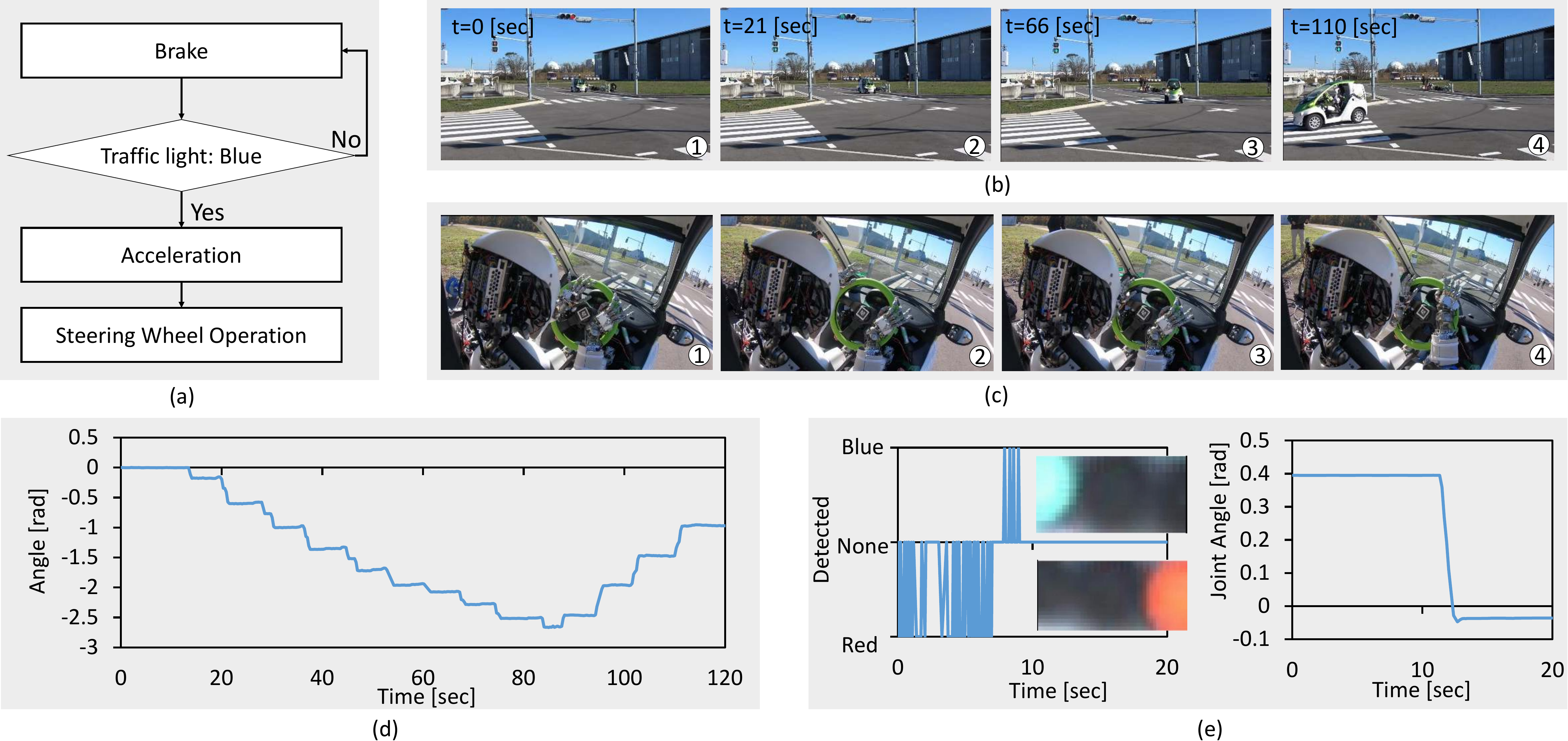}
  \caption{Steering wheel operation with recognition. (a) shows the experimental motion flow. (b) shows the experimental appearance. (c) shows the sequence of steering wheel operation. (d) shows the transition of the steering wheel angle. (e) shows the traffic light recognition result and the joint angle of the left ankle pitch for brake pedal operation.}
  \label{figure:handle-with-recog}
  \vspace{-3.0ex}
\end{figure*}

\subsection{Handle Operation with Recognition} \label{subsec:handle-operation}
\switchlanguage%
{%
  We conducted the steering wheel operation with recognition (\figref{figure:handle-with-recog}, presented in multimedia material).
  We show the motion flow in (a) of \figref{figure:handle-with-recog}.
  The robot steps on the brake pedal when the traffic light is red, releases the brake pedal when the traffic light turns blue, and turns right at the crossing by the steering wheel operation with the static module.
  In this experiment, because turning at a curve is difficult when the car velocity is fast, we use creep velocity just by releasing the brake pedal.
  Also, the steering wheel is turned to the right first and then is turned to the left by a human command.
  (b) shows the overview of this experiment, and we can see that the car began to move when the traffic light turned blue and the car could turn right at the crossing by 90 deg.
  (c) shows the sequence of the steering wheel operation.
  The sequence is rotating the steering wheel by both arms as much as possible, releasing the left hand, returning the left hand to the original position, releasing the right hand, returning the right hand to the original position, and rotating again.
  The transition of the steering angle is shown in (d).
  The robot could turn the steering wheel by about 180 deg in 70 sec.
  The left figure of (e) shows the recognition result of the traffic light.
  The light is cropped by object detection module, and the blue or red of the traffic light is recognized from the ratio of red and blue pixels.
  The module could recognize the moment when the light turns from blue to red.
  At the same time, as shown in the right figure of (e), the left ankle pitch joint moved to release the brake pedal, and the car began to move.
  The problem of this experiment is the slowness of the steering wheel operation.
  Currently, turning at the crossing takes about two minutes, and we must make the motions faster and smoother.
}%
{%
  Handle操作, Recognitionを統合した実験を行った(\figref{figure:handle-with-recog}, presented in multimedia material).
  動作の遷移図を\figref{figure:handle-with-recog}の(a)に示す.
  赤信号のときは止まっており, 青信号になったら発進, static moduleを用いたハンドル操作によって交差点を曲がる.
  本実験では車速が速いと交差点を曲がり切れないため, 車はクリープによって動作し, 信号を認識したらブレーキを無くすことのみを行う.
  また, ハンドルは最初は右に切り, 途中で人間のコマンドによって左に切り始めるようになっている.
  (b)は実験の様子を表しているが, 青信号になったら走り始め, 90度の交差点を曲がりきることができていることがわかる.
  (c)はハンドル操作の様子を表している.
  ハンドル操作のシーケンスは, 両手で回せるところまで回し, 左手を離して元の位置まで戻し, 同様に右手も離して元の位置まで戻し, また両手で回すという一連の動作を繰り返す.
  その際のハンドルの角度遷移は(d)のようになっている.
  約70秒かけて180度近くまでハンドルを回転させることができている.
  (e)の左図は信号認識の様子を表しており, 物体認識によって信号を切り抜き, その中の青の画素と赤の画素の比率で信号の色を認識している.
  赤または青の画素がない場合はNoneとなっており, 信号が赤から青に切り替わる瞬間を読み取れていることがわかる.
  それと同時に, (e)の右図に示すようにブレーキを踏む左足の足首ピッチが戻り, 車が発進している.
  本実験における問題点は, ハンドル操作の遅さである.
  この交差点を曲がるのに約2分間を要しており, 今後クロスハンドルをスムーズに行える方法が必要となると考える.
}%

\section{Limitations and Future Works} \label{sec:discussion}
\switchlanguage%
{%
}%
{%
  本研究では自動運転を題材に, 様々な個々の動作, 認識と行動を統合した運転動作の実験を行った.
  その経験をもとに, 今後の課題を整理する.
}%

\subsection{Pedal Operation}
\switchlanguage%
{%
  The pedal operation is one of the tasks with various remaining issues.
  In this study, we assumed a flat and smooth road and developed a method to achieve the target car velocity quickly by representing the state equation between the car velocity and joint angle of the ankle pitch.
  However, there are three issues.

  First, in the actual driving environment, the road is not smooth; the ground rises at a crossing, and the road is sometimes uphill or downhill.
  In those cases, the state equation trained at a flat road is different from that of the actual environment, and the robot cannot adjust the car velocity well.
  To solve this problem, we need to conduct online learning or add the image of road condition and IMU information in the body to the initial task state $\bm{s}^{task'}$.
  However, online learning becomes difficult with additional network input, and an efficient learning system with only a few data is desired.

  Second, the robot adjusts only the acceleration pedal, and cannot adjust the brake pedal.
  When driving more slowly than creep velocity, the robot must adjust the car velocity by stepping on the brake pedal.
  Also, in this study, although not required at the slow car velocity, the robot needs to acquire how to smoothly step on the brake pedal as the car velocity becomes fast.

  Third, the car velocity is currently obtained from COMS software, but it should be obtained using image information, IMU, etc.
  Since the car velocity obtained from visual odometry is too noisy for pedal operation now, especially at low speed, we need to develop a new estimation algorithm.
}%
{%
  ペダル操作は非常に課題が多い動作の一つである.
  本研究では平面な道路を仮定し, アクセルを踏む角度と車速の状態方程式を記述することで, より素早く目標速度に到達するための手法を開発した.
  しかし, 2つの課題が存在する.

  まず, 実際の運転においては, 舗装が悪く道路が平面でなかったり, 交差点部分で道路が盛り上がっていたり, 上り坂や下り坂があったりする.
  その場合, 平坦な道路で学習させた状態方程式は現実とズレてしまい, 上手く速度調節をすることができない.
  この問題に対しては, オンライン学習を行ったり, 路面状況の画像や身体のIMU等の情報をタスクの初期状態$\bm{s}^{task'}$に含める必要がある.
  しかし, 情報が増えることで学習は困難となり, 今後より少ない試行で学習が可能な機構の開発が望まれる.

  次に, 本研究ではアクセルのみを調節しており, ブレーキを考慮していない.
  実際, クリープ速度よりもゆっくりと走りたい場合, ブレーキで速度を調節することになる.
  また, 本研究では車速が遅いため滑らかにブレーキを踏む必要がなかったが, 車速が今後速くなるにつれ, 滑らかにブレーキを踏む方法も獲得していく必要があると考える.
}%

\subsection{Steering Wheel Operation}
\switchlanguage%
{%
  Although Musashi succeeded in the steering wheel operation with both arms for the first time, there remain some issues.

  First, in this study, the robot operated the steering wheel not by grasping it but by pressing the hand to it.
  Releasing and holding the steering wheel were difficult because the fingers sometimes were caught by the wheel or the robot could not accurately grasp the intended part.
  The first problem is prominent because the complex finger structure is often caught somewhere in the car, and the robot cannot restore itself.
  The hand structure with smooth skin or a glove may be able to solve this problem.
  Also, the robot needs to identify the direction its hand was caught from sensors and restore itself.

  Second, in this study, the robot operated the steering wheel by push-pull steering method, and it is not the usual human motion of the cross arm steering.
  In the cross arm steering, the arms move while interfering with each other, and the inverse kinematics must be solved in a wide range.
  We experienced that large internal force among the arms sometimes emerges and parts of the hands were caught by each other.
  Also, we found that the robot must have the scapula joint to solve the inverse kinematics in a wider range.
}%
{%
  ハンドル操作は, 初めて両手による運転動作を可能としたものの, いくつかの課題が残った.

  まず, 本研究の統合実験ではハンドルを手で握らず, 手を押し付けることでハンドル操作を行っている.
  手でハンドルを握る場合, ハンドルの持ち替えが難しく, 指がどこかに引っかかったり, 正確に意図した場所を握れない等の問題が頻発した.
  これは他の動作においても顕著であり, 指の構造が複雑な車体のどこかに引っかかってしまい, 自力でそれを取ることができなかった.
  引っ掛かりのないような手の構造や, 手袋等の解決策が考えられる.
  また, 引っ掛かりの方向をセンサから同定し, それを自力で戻すような手法の提案も考える必要がある.

  次に, 本研究ではPush Pullと呼ばれる操作によりハンドルを回転させており, いわゆるクロスハンドルは実現できていない.
  クロスハンドルは, 手と手が干渉しながら動作し, かつ非常に広範囲に逆運動学を解く必要がある.
  その際, 手と手の部位の内力や, 部品と筋が引っかかってしまう等の問題が発生した.
  また, 逆運動学を広範囲に解くためには, 肩甲骨関節の必要性が生じることがわかった.
}%

\subsection{Recognition}
\switchlanguage%
{%
  We used the general object detection method for the visual recognition module, and the recognition of the road line and distance to objects are not conducted.
  Especially, the latter is due to the structure of the developed eye unit.
  While the movable eye unit can enlarge the field of view by moving the eyes, the external parameter changes by this movement and estimation of the depth map by stereo vision is difficult.
  Currently, we are developing a learning-based method to estimate the object distance from the convergence angle, binocular parallax, etc.
  Also, our experiments are conducted only at day time.
  At night time, although we, humans, can recognize traffic lights and pedestrians through the cameras, Musashi, that is, YOLOv3, cannot.
  We used the pretrained model of YOLOv3, but the model does not use time-series transition of image and is not trained by using the dataset at the night time.
  Therefore, we need to train another model using video and dataset at the night time.

  Although acoustic information is rarely used for autonomous driving, we applied it to the recognition of the car horn.
  In the future, we need to develop a system for the robot to understand the current situation by integrating the visual and sound information, to recognize the anomaly of the engine, and to talk with humans.

  The recognition of the failure of the hardware components will also become important.
  In musculoskeletal humanoids, the failure is mainly in the muscle, which can be broken when strong force is applied continuously or when the wire is frayed by friction.
  By using an Autoencoder-type network, we can detect anomalies, and even if the muscle is ruptured, by continuing online learning, we can obtain the intersensory network in the broken state as in \cite{kawaharazuka2019longtime}.
  However, the actual task of autonomous driving requires quicker adaptation and operation in dangerous situations, and the detection of anomalies in hardware components and their handling should be considered more deeply for safety.
}%
{%
  視覚による認識には非常に一般的な物体認識手法を用いたが, 現在ではまだ道路のラインの認識や, 物体への距離の認識等は行っていない.
  特に後者は, 目の構造に起因する問題である.
  本研究では可動眼球機構を用いているため, 目を動かすことで視野を広げられる一方, 動かすことで外部パラメータが代わり, ステレオ視によるDepth Mapの推定が難しい.
  現在, 目の角度や視差等の情報から物体までの距離を推定する学習手法を開発中であり, 今後の発展が望まれる.

  また, 聴覚情報はこれまでの自動運転ではほとんど用いられていない情報であるが, 本研究ではcar hornの認識に応用した.
  実際には, 視覚情報と合わせることで音源定位をして状況を理解したり, エンジンの異音の認識や同時に人間と会話できるような能力も身に着けていく必要があると考える.

  認識はこれだけに留まらず, 今後, 身体のハードウェアコンポーネントの故障の認識も重要となると考えられる.
  筋骨格ヒューマノイドにおいて壊れるのは主に筋であり, 強い力がかかり続けたり, 摩擦でワイヤがほつれると切れてしまうことがある.
  Autoencoder型のネットワークを用いることで異常検知が可能となったり, \cite{kawaharazuka2019longtime}のように筋が切れてしまってもオンライン学習を続けることで, 筋が切れた状態における身体感覚を得ることができる.
  しかし, 自動運転というタスクにおいてはより素早い適応や危険時におけるオペレーションが必要であり, 今後安全性のためにハードウェアコンポーネントの異常検知とその対処についても深く考えていく必要がある.
}%

\subsection{Future Works}
\switchlanguage%
{%
  The content we handled in this study is only a part of the entire system for autonomous driving.
  The robot must get into the car, localize the self-position with SLAM, and conduct a driving motion planning.
  Also, the respective components of rotating a key, pulling a handbrake, looking around, acceleration pedal operation, and steering wheel operation must be integrated into one system.
  For this purpose, we need a method that handles the flexible body more accurately, manages the muscle temperature, and recognizes the situation more accurately.
  Also, we need to improve the hardware for the scapula with wide range motion and the body shape with smooth human-like skin.

  Furthermore, in order to be able to drive not only a single car but also various cars, we have to consider what needs to be learned immediately when a car is changed.
  It is necessary to learn not only the current intersensory network and dynamic task control network but also the position of key, side brake, etc. in the future.

  Finally, this is the first example of a robot sitting on a car and driving a car with both arms without a jig, but some metric is needed to compare its performance in the future.
  As for performance, we can think of an index of how long it takes to complete a certain course.
  However, there are many difficult problems such as what kind of course to make and how neatly the course should be driven.
  As for safety, it is more difficult to establish an index.
  Although a long-term endurance test is possible, we have not yet reached the stage.
}%
{%
  本研究で扱った自動運転の内容は, ほんの一部に過ぎない.
  実際には, 車に乗り込んだり, SLAMの情報から自己位置同定, 動作計画等を行う必要がある.
  また, エンジンをかける, サイドブレーキを上げる, 四方の確認, アクセル操作, ハンドル操作などの一つ一つの動作をintegrationしていく必要がある.
  そのためには, 柔軟な身体をより正確に扱う手法の開発, 継続的に動作するための温度管理戦略, より正確な認識機構等が必要になる.
  また, より広い可動域を持つ肩, 人間の皮膚のように引っかかりのない滑らかなフォルム等のハードウェアの改良も今後必要になると考える.

  さらに, 一つの車だけでなく, 様々な車を運転することができるようになるためには, 車が変わった時に何を即座に学習する必要があるのかについて考えなければならない.
  特に, 現状intersensory networkとdynamic task control networkが車が変わった際に学習すべきcomponentであるが, 今後, キーやサイドブレーキの位置なども学習する必要があると考える.

  最後に, ジグ等無しにロボットが車に座り両手で実際の車を運転した例は初であるが, この先そのパフォーマンスを比較するうえでは何らかのメトリックが必要となる.
  パフォーマンスとしては, あるコースをどのくらいの時間で走り切るか, という指標が考えられるが, どのようなコースにするかやどの程度綺麗に走れば良いのか等難しい問題が山積みである.
  また, 安全性に関してはより指標の策定が困難であり, 長時間の耐久試験等が考えられるが, 現状その段階までは至らない.
}%

\section{CONCLUSION} \label{sec:conclusion}
\switchlanguage%
{%
  In this study, we focused on autonomous driving by the musculoskeletal humanoid Musashi using the characteristics of its hardware and software.
  By making use of the flexibility, variable stiffness structure, and several sensors, we succeeded in the steering wheel operation with both arms and human recognition in the side mirror.
  Also, we proposed a learning-based system handling the flexible body with difficult modeling and succeeded in the pedal and steering wheel operations with recognition.
  For autonomous driving by humanoids in the future, we would like to develop the next hardware and software using the obtained knowledge.
}%
{%
  本研究では, 筋骨格ヒューマノイドによる自動運転に焦点を当て, そのハードウェアとソフトウェアの特性を用いた様々な運転操作を行った.
  柔軟かつ, その柔軟さを変化させることができる点, センシングに優れる点を活かし, 両手による運転やサイドミラー認識等も成功させている.
  また, その柔軟でモデル化困難な身体を扱うための学習型システム構成を提案し, それをもとに認識を含むペダル操作やハンドル操作を可能とした.
  今後も, 未来のヒューマノイドにおける自動運転のために, 得られた知見を活かしてハードウェア・ソフトウェアの開発に励みたい.
}%

{
  \bibliographystyle{IEEEtran}
  \bibliography{main}

\begin{thebibliography}{10}
\providecommand{\url}[1]{#1}
\csname url@rmstyle\endcsname
\providecommand{\newblock}{\relax}
\providecommand{\bibinfo}[2]{#2}
\providecommand\BIBentrySTDinterwordspacing{\spaceskip=0pt\relax}
\providecommand\BIBentryALTinterwordstretchfactor{4}
\providecommand\BIBentryALTinterwordspacing{\spaceskip=\fontdimen2\font plus
\BIBentryALTinterwordstretchfactor\fontdimen3\font minus
  \fontdimen4\font\relax}
\providecommand\BIBforeignlanguage[2]{{%
\expandafter\ifx\csname l@#1\endcsname\relax
\typeout{** WARNING: IEEEtran.bst: No hyphenation pattern has been}%
\typeout{** loaded for the language `#1'. Using the pattern for}%
\typeout{** the default language instead.}%
\else
\language=\csname l@#1\endcsname
\fi
#2}}

\bibitem{levinson2011driving}
J.~Levinson, J.~Askeland, J.~Becker, J.~Dolson, D.~Held, S.~Kammel, J.~Z.
  Kolter, D.~Langer, O.~Pink, V.~Pratt, M.~Sokolsky, G.~Stanek, D.~Stavens,
  A.~Teichman, M.~Werling, and S.~Thrun, ``{Towards fully autonomous driving:
  Systems and algorithms},'' in \emph{Proceedings of the 2011 IEEE Intelligent
  Vehicles Symposium (IV)}, 2011, pp. 163--168.

\bibitem{endsley2017tesla}
M.~R. Endsley, ``{Autonomous Driving Systems: A Preliminary Naturalistic Study
  of the Tesla Model S},'' \emph{Journal of Cognitive Engineering and Decision
  Making}, vol.~11, no.~3, pp. 225--238, 2017.

\bibitem{darpa2015drc}
``{DARPA Robotics Challenge},''
  \url{http://archive.darpa.mil/roboticschallenge/}, {Accessed: 2021-06-30}.

\bibitem{rasmussen2014driving}
C.~Rasmussen, K.~Sohn, Q.~Wang, and P.~Oh, ``{Perception and control strategies
  for driving utility vehicles with a humanoid robot},'' in \emph{Proceedings
  of the 2014 IEEE/RSJ International Conference on Intelligent Robots and
  Systems}, 2014, pp. 973--980.

\bibitem{paolillo2018driving}
A.~Paolillo, P.~Gergondet, A.~Cherubini, M.~Vendittelli, and A.~Kheddar,
  ``{Autonomous car driving by a humanoid robot},'' \emph{Journal of Field
  Robotics}, vol.~35, no.~2, pp. 169--186, 2018.

\bibitem{gravato2010ecce1}
H.~G. Marques, M.~J{\"a}ntsh, S.~Wittmeier, O.~Holland, C.~Alessandro,
  A.~Diamond, M.~Lungarella, and R.~Knight, ``{ECCE1: the first of a series of
  anthropomimetic musculoskeletal upper torsos},'' in \emph{Proceedings of the
  2010 IEEE-RAS International Conference on Humanoid Robots}, 2010, pp.
  391--396.

\bibitem{nakanishi2013design}
Y.~Nakanishi, S.~Ohta, T.~Shirai, Y.~Asano, T.~Kozuki, Y.~Kakehashi,
  H.~Mizoguchi, T.~Kurotobi, Y.~Motegi, K.~Sasabuchi, J.~Urata, K.~Okada,
  I.~Mizuuchi, and M.~Inaba, ``{Design Approach of Biologically-Inspired
  Musculoskeletal Humanoids},'' \emph{International Journal of Advanced Robotic
  Systems}, vol.~10, no.~4, pp. 216--228, 2013.

\bibitem{asano2016kengoro}
Y.~Asano, T.~Kozuki, S.~Ookubo, M.~Kawamura, S.~Nakashima, T.~Katayama,
  Y.~Iori, H.~Toshinori, K.~Kawaharazuka, S.~Makino, Y.~Kakiuchi, K.~Okada, and
  M.~Inaba, ``{Human Mimetic Musculoskeletal Humanoid Kengoro toward Real World
  Physically Interactive Actions},'' in \emph{Proceedings of the 2016 IEEE-RAS
  International Conference on Humanoid Robots}, 2016, pp. 876--883.

\bibitem{kawaharazuka2019musashi}
K.~Kawaharazuka, S.~Makino, K.~Tsuzuki, M.~Onitsuka, Y.~Nagamatsu, K.~Shinjo,
  T.~Makabe, Y.~Asano, K.~Okada, K.~Kawasaki, and M.~Inaba, ``{Component
  Modularized Design of Musculoskeletal Humanoid Platform Musashi to
  Investigate Learning Control Systems},'' in \emph{Proceedings of the 2019
  IEEE/RSJ International Conference on Intelligent Robots and Systems}, 2019,
  pp. 7294--7301.

\bibitem{haug2004crash}
E.~Haug, H.~Choi, S.~Robin, and M.~Beaugonin, ``{Human Models for Crash and
  Impact Simulation},'' in \emph{Computational Models for the Human Body}, ser.
  Handbook of Numerical Analysis.\hskip 1em plus 0.5em minus 0.4em\relax
  Elsevier, 2004, vol.~12, pp. 231--452.

\bibitem{makabe2018eyeunit}
T.~Makabe, K.~Kawaharazuka, K.~Tsuzuki, K.~Wada, S.~Makino, M.~Kawamura,
  A.~Fujii, M.~Onitsuka, Y.~Asano, K.~Okada, K.~Kawasaki, and M.~Inaba,
  ``{Development of Movable Binocular High-Resolution Eye-Camera Unit for
  Humanoid and the Evaluation of Looking Around Fixation Control and Object
  Recognition},'' in \emph{Proceedings of the 2018 IEEE-RAS International
  Conference on Humanoid Robots}, 2018, pp. 840--845.

\bibitem{makino2018hand}
S.~Makino, K.~Kawaharazuka, M.~Kawamura, A.~Fujii, T.~Makabe, M.~Onitsuka,
  Y.~Asano, K.~Okada, K.~Kawasaki, and M.~Inaba, ``{Five-Fingered Hand with
  Wide Range of Thumb Using Combination of Machined Springs and Variable
  Stiffness Joints},'' in \emph{Proceedings of the 2018 IEEE/RSJ International
  Conference on Intelligent Robots and Systems}, 2018, pp. 4562--4567.

\bibitem{shinjo2019foot}
K.~Shinjo, K.~Kawaharazuka, Y.~Asano, S.~Nakashima, S.~Makino, M.~Onitsuka,
  K.~Tsuzuki, K.~Okada, K.~Kawasaki, and M.~Inaba, ``{Foot with a Core-shell
  Structural Six-axis Force Sensor for Pedal Depressing and Recovering from
  Foot Slipping during Pedal Pushing Toward Autonomous Driving by Humanoids},''
  in \emph{Proceedings of the 2019 IEEE/RSJ International Conference on
  Intelligent Robots and Systems}, 2019, pp. 3049--3054.

\bibitem{kawaharazuka2019longtime}
K.~Kawaharazuka, K.~Tsuzuki, S.~Makino, M.~Onitsuka, Y.~Asano, K.~Okada,
  K.~Kawasaki, and M.~Inaba, ``{Long-time Self-body Image Acquisition and its
  Application to the Control of Musculoskeletal Structures},'' \emph{IEEE
  Robotics and Automation Letters}, vol.~4, no.~3, pp. 2965--2972, 2019.

\bibitem{kawaharazuka2019pedal}
K.~Kawaharazuka, K.~Tsuzuki, S.~Makino, M.~Onitsuka, K.~Shinjo, Y.~Asano,
  K.~Okada, K.~Kawasaki, and M.~Inaba, ``{Task-specific Self-body Controller
  Acquisition by Musculoskeletal Humanoids: Application to Pedal Control in
  Autonomous Driving},'' in \emph{Proceedings of the 2019 IEEE/RSJ
  International Conference on Intelligent Robots and Systems}, 2019, pp.
  813--818.

\bibitem{kawaharazuka2019relax}
K.~Kawaharazuka, K.~Tsuzuki, M.~Onitsuka, Y.~Koga, Y.~Omura, Y.~Asano,
  K.~Okada, K.~Kawasaki, and M.~Inaba, ``{Reflex-based Motion Strategy of
  Musculoskeletal Humanoids under Environmental Contact Using Muscle Relaxation
  Control},'' in \emph{Proceedings of the 2019 IEEE-RAS International
  Conference on Humanoid Robots}, 2019, pp. 114--119.

\bibitem{asano2015sensordriver}
Y.~Asano, T.~Kozuki, S.~Ookubo, K.~Kawasaki, T.~Shirai, K.~Kimura, K.~Okada,
  and M.~Inaba, ``{A Sensor-driver Integrated Muscle Module with High-tension
  Measurability and Flexibility for Tendon-driven Robots},'' in
  \emph{Proceedings of the 2015 IEEE/RSJ International Conference on
  Intelligent Robots and Systems}, 2015, pp. 5960--5965.

\bibitem{kawaharazuka2017forearm}
K.~Kawaharazuka, S.~Makino, M.~Kawamura, Y.~Asano, Y.~Kakiuchi, K.~Okada, and
  M.~Inaba, ``{Human Mimetic Forearm Design with Radioulnar Joint using
  Miniature Bone-muscle Modules and its Applications},'' in \emph{Proceedings
  of the 2017 IEEE/RSJ International Conference on Intelligent Robots and
  Systems}, 2017, pp. 4956--4962.

\bibitem{rumelhart1986backprop}
D.~E. Rumelhart, G.~E. Hinton, and R.~J. Williams, ``{Learning representations
  by back-propagating errors},'' \emph{nature}, vol. 323, no. 6088, pp.
  533--536, 1986.

\bibitem{redmon2018yolov3}
J.~Redmon and A.~Farhadi, ``{YOLOv3: An Incremental Improvement},'' arXiv
  preprint arXiv:1804.02767, 2018.

\end{thebibliography}
}

\end{document}